\documentclass[pdflatex, sn-basic]{sn-jnl}
\usepackage{graphicx}
\usepackage{booktabs} 
\usepackage{caption} 
\PassOptionsToPackage{hyphens}{url}\usepackage{hyperref}
\usepackage{array} 
\newcolumntype{V}[1]{>{\raggedright\arraybackslash}p{#1}} 

\usepackage{mathptmx}
\usepackage{amsmath}
\usepackage{lineno}
\usepackage{balance}
\newcommand{\bi}{\begin{itemize}}
\newcommand{\ei}{\end{itemize}}
\renewcommand{\footnotesize}{\scriptsize}
\usepackage{booktabs}
\usepackage[ruled,vlined]{algorithm2e}

\begin{document}
\title{Combating Toxic Language: A Review of LLM-Based Strategies for Software Engineering}

\author[1]{\fnm{Hao} \sur{Zhuo}}\email{hz324@cornell.edu}\equalcont{These authors contributed equally to this work.}

\author[1]{\fnm{Yicheng} \sur{Yang}}\email{yy546@cornell.edu}
\equalcont{These authors contributed equally to this work.}

\author*[2]{\fnm{Kewen} \sur{Peng}}\email{kpeng@ncsu.edu}

\affil*[1]{\orgdiv{Bowers College of Computing and Information Science}, \orgname{Cornell University}, \orgaddress{\street{410 Thurston Avenue}, \city{Ithaca}, \postcode{14850}, \state{NY}, \country{United States}}}

\affil[2]{\orgdiv{Department of Computer Science}, \orgname{North Carolina State University}, \orgaddress{\street{2101 Hillsborough Street}, \city{Raleigh}, \postcode{27695}, \state{NC}, \country{United States}}}

\abstract{Large Language Models (LLMs) have become integral to Software Engineering (SE), increasingly used in development workflows. However, their widespread adoption raises concerns about the presence and propagation of toxic language—harmful or offensive content that can foster exclusionary environments. This paper provides a comprehensive review of recent research (2020-2024) on toxicity detection and mitigation, focusing on both SE-specific and general-purpose datasets. We examine annotation and pre-processing techniques, assess detection methodologies, and evaluate mitigation strategies, particularly those leveraging LLMs. Additionally, we conduct an ablation study demonstrating the effectiveness of LLM-based rewriting for reducing toxicity. This review is limited to studies published within the specified timeframe and within the domain of toxicity in LLMs and SE; therefore, certain emerging methods or datasets beyond this period may fall outside its purview. By synthesizing existing work and identifying open challenges, this review highlights key areas for future research to ensure the responsible deployment of LLMs in SE and beyond.}


\maketitle

\section{Introduction} 
In Software Engineering (SE), toxic language within developer communications can hamper collaboration and productivity. Indeed, the increasing integration of Large Language Models (LLMs), such as BERT and GPT-4, into SE workflows, alongside their extensive applications across diverse domains like education, research, healthcare, and finance~\citep{dam2024complete}, has amplified this concern. For instance, while LLM-powered chatbots and virtual assistants are transforming customer support by offering efficient and personalized services~\citep{pandya2023automating}, and in the healthcare sector, LLMs are contributing to advancements in medical research, drug discovery, and patient care~\citep{athota2020chatbot}, their widespread use has drawn increasing attention to their societal implications, both positive and negative.

One pressing concern in the digital age is toxic language, which refers to harmful, offensive, or discriminatory language. The prevalence of such toxic language on online platforms, such as social media and forums, creates hostile and exclusionary environments that hinder constructive dialogue and can cause significant harm to individuals~\citep{mohan2017impact}. While commonly observed in general online discourse, toxicity in SE often manifests uniquely within technical contexts like code review discussions, issue trackers, and pull request comments, characterized by a blend of technical jargon and collaborative discourse. Researchers actively develop detection and mitigation methods, yet challenges persist in accurately identifying toxic language across diverse contexts and languages. For example,~\citet{radfar2020characterizing} found that mildly offensive terms are more frequently used in tweets to express hostility between users with no social connection than between users who are mutual friends.

LLMs, with their ability to process and analyze vast amounts of textual data, offer promising solutions for addressing these challenges~\citep{cheriyan2021towards, mishra2024exploring}. Unlike traditional methods, LLMs can detect nuanced patterns of harmful language, including subtle and context-dependent forms of toxicity, that might otherwise go unnoticed. Researchers have also proposed and implemented various strategies to detect and mitigate toxicity in LLM-generated content, underscoring the potential of LLMs to play a central role in promoting healthier online interactions.

This work synthesizes insights from a substantial body of publications on toxicity research conducted from 2020 to 2024. It delves into core topics such as toxicity detection and mitigation while also addressing critical auxiliary processes like corpus selection and data pre-processing. By consolidating findings from diverse studies, this paper provides a comprehensive overview of the current state of research and highlights gaps that require further exploration. One significant challenge with LLMs is their potential to generate toxic content, even in the absence of explicit prompts encouraging such behavior~\citep{gehman2020realtoxicityprompts}. This issue stems from factors such as the presence of toxic content in training data~\citep{gpt2knowyournumber}, inherent algorithmic biases, or malicious exploitation. To better understand and address these issues, this paper proposes future research directions, including an ablation study that sheds light on the mechanisms underlying these behaviors and explores strategies for improvement.

Our main contributions include:
\begin{itemize}
    \item We systematically compare SE and non-SE datasets in the context of toxicity research.
    \item We categorize and summarize common annotation and pre-processing steps.
    \item We evaluate toxicity detection and mitigation solutions.
    \item We recommend using LLM-based rewriting approaches for toxicity mitigation, based on the results of our ablation study.
\end{itemize}

While existing literature includes comprehensive surveys on offensive language and hate speech detection~\citep{mansur2023twitter, alkomah2022literature, jahan2023systematic}, which primarily focus on general online discourse and algorithmic detection methods, more recent reviews have begun to specifically address toxicity in LLMs, encompassing both detection and mitigation strategies~\citep{villate2024systematic}. Our work, however, distinguishes itself through several key aspects:
\begin{itemize}
    \item Unique focus on toxicity in SE communications, addressing its specific contextual nuances.
    \item Examination of challenges from LLM-generated toxicity and their mitigation role, including experiment-backed insights from an ablation study on LLM-based rewriting approaches—a novel contribution not found in existing surveys.
    \item Detailed analysis of critical auxiliary processes: data annotation and pre-processing.
    \item Consideration of ethical and policy implications for toxicity detection system deployment, an area less explored in existing reviews.
\end{itemize}

The remainder of this paper is organized as follows:
\S\ref{background} introduces the use of LLM techniques in the SE domain, with a focus on toxicity analysis.
\S\ref{methodology} outlines common techniques for annotating, pre-processing, and utilizing datasets for toxicity detection and mitigation.
\S\ref{future} outlines improvements in pre-processing and toxicity detection, alongside recommendations for ethical governance.
\S\ref{ablation} details the setup, methodology, and results of our ablation study.
The overall workflow is illustrated in Figure~\ref{fig:Overall_Workflow}.
Our code is available in an online repository: \url{https://github.com/yangjack1998/SE-Mitigate}

\begin{figure*}[htp]
    \centering
    \includegraphics[width=\textwidth]{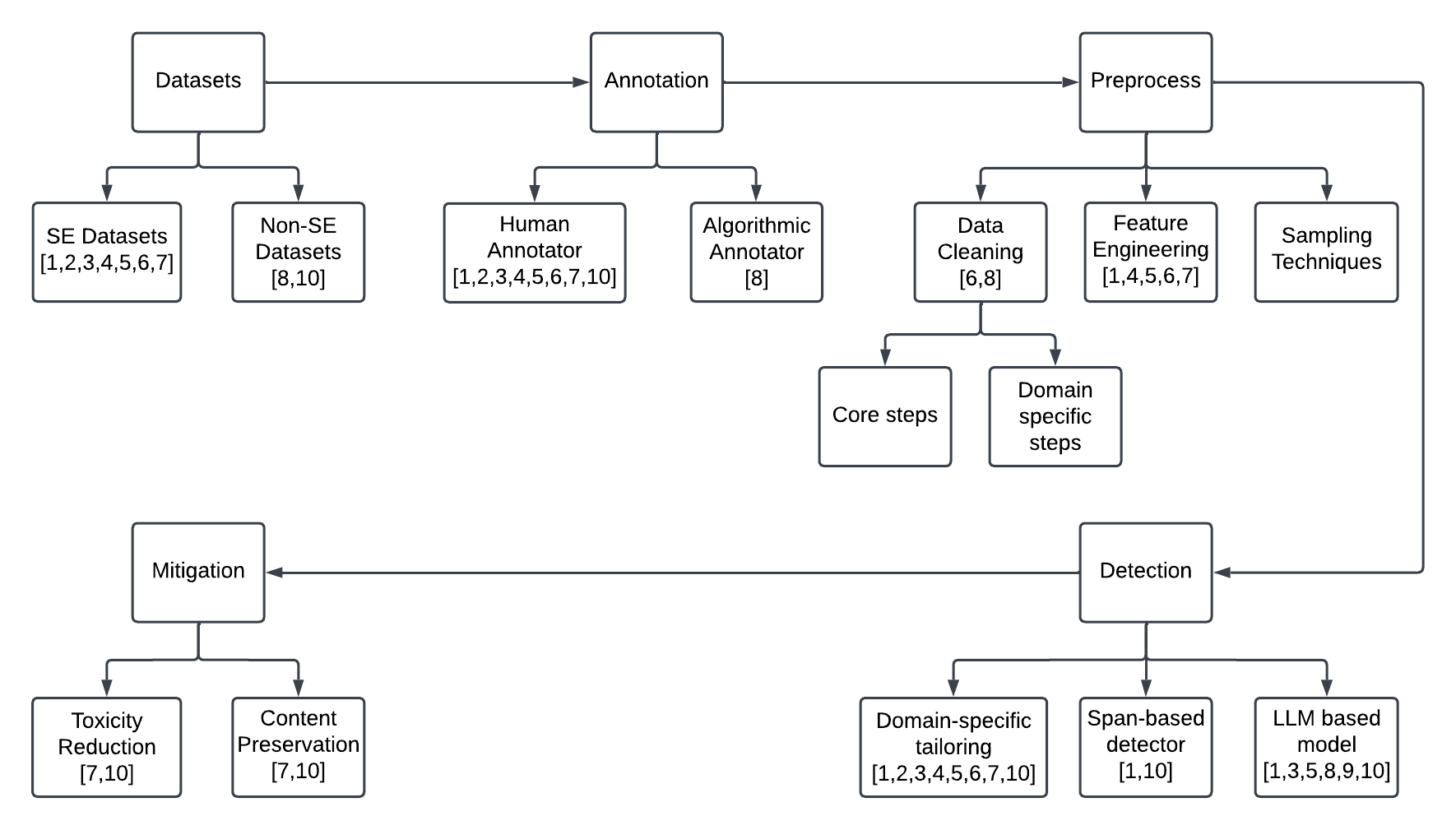}
    \caption{Overall workflow of LLM toxicity detection and mitigation. Number annotations in each step represent relevant core literature as indexed in Table~\ref{tab:core-literature}.
    }
    \label{fig:Overall_Workflow}
\end{figure*}

\section{Background}\label{background}
\subsection{Why is LLM in SE important}
The integration of LLMs into SE has revolutionized the field, offering substantial improvements in automation, efficiency, and decision-making. From code generation and summarization to commit classification and toxicity detection, LLMs are being increasingly leveraged to enhance software development workflows. This transformative impact stems from their ability to effectively process and understand both natural language descriptions and programming code, a capability exemplified by models like CodeBERT~\citep{feng2020codebert}.

Many SE tasks can naturally be formulated as data analysis (learning) tasks such as classification tasks,  regression tasks, and generative tasks~\citep{wang2022machine}. These tasks are \textbf{well-suited to being addressed by LLMs} and hold significant potential for optimization and refinement through their application. Specifically, the underlying transformer architectures~\citep{vaswani2023attentionneed} of modern LLMs, with their remarkable ability to model long-range dependencies and capture intricate contextual relationships within sequences, are particularly well-suited to the complexities of SE data. 

For instance, automated code summarization, where the objective is to generate concise and descriptive natural language summaries for given code snippets or functions, can be naturally cast as a generative task. ~\citet{zhang2020retrieval} propose a retrieval-based neural model to aid code summarization that improves the state-of-the-art methods by enhancing the model with the most similar code snippets retrieved from the training set. Unlike traditional recurrent neural networks (RNNs) that can struggle with very long sequences common in code or extended developer discussions, transformers effectively process these lengthy contexts~\citep{vaswani2023attentionneed}, making them highly advantageous for SE-specific tasks involving technical vocabulary and structured code elements.

~\citet{hou2024large} point out that among their collection of LLM studies for SE tasks, following generation tasks, which the majority (about 70.97\%) of studies center around, there are classification tasks that take up 21.62\%. For example, categorizing commits into maintenance tasks helps practitioners make informed decisions regarding resource allocation and effectively manage technical debt. Researchers continue to explore ways to enhance the accuracy of this task through the application of LLMs. A study~\citep{ghadhab2021augmenting} proposed leveraging BERT (Bidirectional Encoder Representations from Transformers) to classify commits into three categories of maintenance tasks: corrective, perfective, and adaptive. Their approach demonstrated an approximately 8\% improvement in accuracy compared to the previous state-of-the-art model.

In addition to the direct automation of coding tasks, LLMs are also poised for \textbf{benefiting the SE community} itself by addressing social challenges. Toxic interaction can disrupt teamwork, discourage participation, and reduce productivity in software projects.~\citet{raman2020stress} take an early step in implementing and illustrating a measurement instrument (an SVM classifier tailored for software engineering) to detect toxic discussions in GitHub issues. They also recognize the limited availability of large, software engineering-specific toxicity datasets, as well as the significant class imbalance they often exhibit (e.g., only 6 out of 1000 GitHub issues are toxic). To address this issue,~\citet{mishra2024exploring} explore the use of a zero-shot LLM (ChatGPT) for detecting toxicity in software-related text. Their findings show promising accuracy, underscoring the potential for integrating ChatGPT-enabled toxicity detection into developer communication channels. Currently, there is a lack of studies that go further to explore the mitigation of toxicity in software-related text using LLMs. To address this gap, an ablation study will be presented later in the paper to demonstrate the feasibility of such an approach.

In terms of the distribution of LLM usages in SE activities, among six phases of the Software Development Life Cycle (SDLC) (i.e., requirements engineering, software design, software development, software quality assurance, software maintenance, and software management), over half of the studies are observed in the software development domain~\citep{hou2024large}. Within the domain, code generation and code completion are two tasks of great interest. While LLMs' assistance with these tasks could largely improve the efficiency of SE practitioners, they have more subtle implications that we should be aware of. An empirical study~\citep{gehman2020realtoxicityprompts} finds that pretrained neural language models (LMs) are susceptible to producing racist, sexist, or other forms of toxic language, posing challenges to their safe deployment. The tendency of pretrained LMs to generate toxic text even from seemingly harmless prompts highlights the importance of improving data selection processes during pretraining. This concern extends to similar generative processes, including code generation, as they are fundamentally driven by LLMs trained for text generation tasks.

\subsection{What is toxicity}
\textbf{Understanding and defining toxicity} is crucial for fostering healthy online and professional interactions within software engineering. Toxicity is a broad and more subjective challenge compared to problems like spam detection~\citep{sarker2022identification}. The Perspective API~\citep{conversationAI} defines toxic content as ``texts that are rude, disrespectful, or unreasonable". ~\citet{gehman2020realtoxicityprompts} put more focus on the harm it creates, and describe toxic language as any form of speech or text that includes hate speech, offensive comments, harassment, or abusive behavior intended to harm, insult, or threaten others. A prior study~\citep{villate2024systematic} observed specific characteristics of toxicity definitions based on the domain of study. For the purpose of this paper and in alignment with the professional nature of the SE domain, we adopt the comprehensive definition provided by~\citet{sarker2020benchmark}: ``An SE conversation will be considered as toxic, if it includes any of the following: i) offensive name calling, ii) insults, iii) threats, iv) personal attacks, v) flirtations, vi) reference to sexual activities, and vii) swearing or cursing."

\textbf{Annotation guidelines} play a crucial role in ensuring consistency in toxicity detection datasets. Annotators follow standardized instructions to label instances of toxicity, helping to train models with reliable data. For example, the models powering certain toxicity detection systems, such as the Perspective API, rely on manual annotations guided by published annotation guidelines~\citep{annotation2018}. However, while clear and well-defined guidelines improve annotation accuracy and consistency, they do not eliminate subjectivity. Biases in guideline design, cultural differences among annotators, and varying interpretations of toxicity can introduce inconsistencies that affect the generalization of the model in different contexts~\citep{fairchild2020interrater}. A critical challenge in toxicity research is determining whether guidelines strike the right balance between recall and precision: Overly broad definitions risk flagging benign language, while narrow definitions may fail to capture implicit toxicity.

To address these challenges in practical settings, particularly within software engineering, detailed annotation rubrics are employed to guide human annotators. For instance,~\citet{sarker2020benchmark} empirically developed a rubric for classifying toxicity in SE conversations, which includes specific rules for identifying the following types of toxic content:
\begin{itemize}
\item Profanity or curse words: Direct use of profane or vulgar language.
\item Acronyms referring to expletives: Initialisms or abbreviations that stand for expletives.
\item Insults towards a person or their work: Direct insults or demeaning remarks aimed at an individual or their professional contributions.
\item Identity attacks: Attacks targeting an individual's identity, such as those based on race, gender, religion, or other protected characteristics.
\item Threats: Explicit or implicit threats of harm, intimidation, or violence.
\item References to sexual activities: Explicit or implicit mentions of sexual activities.
\item Flirtations: Unwelcome flirtatious comments or sexual advances.
\end{itemize}
Crucially, this rubric also distinguishes self-deprecating language (e.g., using words like ``dumb" or ``stupid" when referring to oneself or one's own work) as non-toxic, recognizing its common use in SE communities for expressing mistakes without malicious intent. These rules provide concrete examples and decision-making criteria, helping annotators to accurately distinguish between technical disagreements and genuinely toxic interactions, thereby ensuring that collected data accurately reflects the nuances of toxicity within SE communications.

Despite comprehensive rubrics, the subjective nature of toxicity often leads to ambiguities in real-world contexts. Consider the following comment~\citep{raman2020stress}: \textit{``Let's remove anti-cheat motion, which delivers more of a pain in the ass. After all, it is also possible to add a plugin."} While the phrase \textit{``pain in the ass"} contains profanity that might be flagged as toxic by some guidelines, its use here could be interpreted by others as a colloquial, non-malicious expression of difficulty within a technical context. Such borderline cases highlight the inherent difficulties in achieving perfect consensus during annotation.

\textbf{Inter-annotator agreement (IAA)} is a key metric to evaluate the consistency and reliability of annotations. Given the subjective nature of toxic language, ensuring high agreement is essential to produce reliable data sets~\citep{sarker2022identification, sarker2020benchmark, sarker2023toxispanse}. High IAA indicates that annotation guidelines are effective and that annotators have a shared understanding of toxicity. Among the widely used methods for assessing IAA, Cohen’s $\kappa$~\citep{cohen1960coefficient} accounts for agreement occurring by chance and is calculated as $\kappa = \frac{P_o - P_e}{1 - P_e}$, where $P_o$ represents observed agreement and $P_e$ represents expected agreement by chance. In the ``pain in the ass" example, if one annotator labels it as toxic ($A_T$) and another as non-toxic ($A_NT$), this disagreement directly impacts the observed agreement ($P_0$) and consequently lowers the Kappa score, signaling a need to refine guidelines or provide more context. Another commonly used metric, Krippendorff’s $\alpha$~\citep{krippendorff2018content}, generalizes inter-rater reliability by accommodating multiple annotators, diverse data types, and missing data. It is computed as $\alpha = 1 - \frac{D_o}{D_e}$, where $D_o$ represents observed disagreement and $D_e$ represents expected disagreement by chance. These measures are critical for validating the robustness and consistency of annotations in toxicity detection research. Yet, they have limitations—low agreement may indicate annotator disagreement, but it may also reveal deeper issues such as ambiguous labeling guidelines~\citep{sandri2023don} or the need for more context-aware annotation strategies.

\textbf{Model performance} is another essential aspect used to assess the effectiveness of toxicity detection systems, beyond annotation reliability. Performance metrics such as precision, recall, accuracy, F1 score, and F0.5 score provide insights into different aspects of model behavior. These metrics ensure a comprehensive understanding of how well a model balances false positives and false negatives. Table~\ref{tab:confusion-matrix} presents the confusion matrix, which serves as the foundation for calculating these metrics, while Table~\ref{tab:performance-metrics} defines and details their respective equations. By employing rigorous evaluation metrics, researchers can gauge the strengths and limitations of toxicity detection models and refine them for improved real-world application. These metrics are critically used in our ablation study (\S\ref{ablation}) to evaluate and compare different aspects of model performance.

\begin{table}[h!]
\small
\centering
\caption{Confusion matrix in binary classification}
\begin{tabular}{ccc}
\hline
\toprule
    \textbf{} & \textbf{Predicted Positive} & \textbf{Predicted Negative} \\
\midrule
    \textbf{Actual Positive} & True Positive (TP) & False Negative (FN) \\
\midrule
    \textbf{Actual Negative} & False Positive (FP) & True Negative (TN) \\
\bottomrule
\end{tabular}
\label{tab:confusion-matrix}
\end{table}

\begin{table*}[h!]
\centering
\caption{Model performance metrics in binary classification}
\resizebox{\textwidth}{!}{%
\begin{tabular}{ccc}
\hline
\toprule
    \textbf{Metric} & \textbf{Definition} & \textbf{Equation} \\ 
\midrule
    Precision & 
    Proportion of correctly predicted positive instances out of all instances predicted as positive & 
    \(\frac{\text{TP}}{\text{TP} + \text{FP}}\) \\
    \hline
    Recall & 
    Proportion of correctly predicted positive instances out of all actual positive instances & 
    \(\frac{\text{TP}}{\text{TP} + \text{FN}}\) \\ 
    \hline
    Accuracy & 
    Proportion of correctly predicted instances (both positive and negative) out of all instances & 
    \(\frac{\text{TP} + \text{TN}}{\text{TP+FP+TN+FN}}\) \\ 
    \hline
    F1 Score & 
    Harmonic mean of precision and recall, balancing the trade-off between the two metrics & 
    \(2 \cdot \frac{\text{Precision} \cdot \text{Recall}}{\text{Precision} + \text{Recall}}\) \\ \hline
    F0.5 Score & 
    Variation of F1 score that places more emphasis on precision than recall & 
    \((1 + 0.5^2) \cdot \frac{\text{Precision} \cdot \text{Recall}}{0.5^2 \cdot \text{Precision} + \text{Recall}}\) \\ 
\bottomrule
\end{tabular}%
}
\label{tab:performance-metrics}
\end{table*}

\subsection{Empirical corpus}
Given toxicity's diverse nature, from offensive language to harmful behavior, datasets must be specifically tailored. This section provides a comprehensive overview of toxicity datasets, focusing on both SE and non-SE contexts. Distinguishing these domains is crucial for understanding their unique challenges and characteristics: SE datasets such as~\citet{raman2020stress} and~\citet{sarker2023toxispanse}, from developer communication platforms, present challenges like technical jargon and collaborative conversations, while non-SE datasets such as~\citet{gehman2020realtoxicityprompts} and~\citet{xu2022leashing} often feature broader forms of toxicity, including hate speech and cyberbullying. This segmentation allows researchers to identify common trends, evaluate cross-domain model applicability, and highlight areas for future dataset development.

However, existing datasets, particularly in the SE domain, often present notable limitations regarding representativeness and sampling biases. A pervasive issue is significant class imbalance, with toxic instances being rare; for example, the ToxiCR dataset contains only 4.4\% toxic instances~\citep{sarker2023automated}. This extreme rarity makes it unfeasible to gather sufficient toxic examples via random sampling. Consequently, researchers must employ specific, often biased, sampling strategies. While efforts like combining multiple detection methods (e.g., language-based detectors, locked issues) to enhance sample diversity~\citep{miller2022did} are crucial for data collection, they inherently risk biasing the type of toxicity captured. This known trade-off between sample size and representativeness limits the broad applicability of resulting datasets and can cause models to struggle in effectively detecting the minority (toxic) class, thereby restricting their practical utility. This dataset imbalance can be addressed in future work by leveraging over-sampling or down-sampling methods to synthesize minority labels, such as SMOTE (Synthetic Minority Over-sampling Technique)~\citep{chawla2002smote}, which creates synthetic examples for the minority class by interpolating between existing minority instances rather than simply duplicating them.

Furthermore, current SE toxicity datasets are overwhelmingly English-centric, an oversight given the global, multilingual nature of software development. While multilingual data have been utilized to study toxicity in general online discourse~\citep{leite2020toxiclanguagedetectionsocial}, its application in SE has been less prominent. This English predominance, with limited non-English SE contexts (e.g., Chinese or Arabic developer forums), suggests a linguistic bias potentially affecting research generalizability. Thus, curating multilingual SE toxicity data is crucial for broader coverage and more inclusive detection systems worldwide.

Tables~\ref{tab:toxicity-se-datasets} and~\ref{tab:toxicity-non-se-datasets} provide a summary of key SE and non-SE datasets used in toxicity detection research, organized by their characteristics, sources, annotation methods, and evaluation metrics. This overview provides a foundational reference for researchers seeking to build on existing work, compare approaches, and identify gaps in dataset development to enhance toxicity detection systems.

\subsection{Study selection}
To identify relevant articles, we conducted a systematic search across four major academic databases: IEEE Xplore, ACM Digital Library, Springer, and the preprint server, arXiv. The inclusion of arXiv is crucial for capturing the most recent, innovative, and early-stage findings in the dynamic LLM-toxicity intersection before formal peer review. This practice aligns with best practices for multivocal literature reviews (MLRs) in software engineering, where incorporating grey literature like preprints is recommended to capture current knowledge~\citep{garousi2019guidelines}. The risk associated with including unreviewed preprints is mitigated by a subsequent rigorous screening process (detailed below). Given our focus on recent software engineering advancements in toxicity, we restricted our search to articles published between 2020 and 2024.

Our initial search query, which utilized the core terms \textit{toxic} and \textit{large language models}, yielded an initial pool of 2,280 studies. The selection of these keywords was systematically motivated by their effectiveness in maximizing recall. We performed initial pilot searches to scope the vocabulary used in this domain, specifically comparing the noun \textit{toxicity} with the adjective \textit{toxic}. The latter provided significantly broader coverage in the target databases and was thus selected as the primary identifier for this dimension of the search to strategically prioritize recall over precision. The specific queries used for each of the libraries are detailed in Table~\ref{tab:library-queries}.

The selection of primary studies followed a structured, multi-stage screening method to ensure the relevance, quality, and uniqueness of the final study set, starting from an initial pool of 2,280 studies. This general process adheres to the well-established guidelines for conducting Systematic Literature Reviews (SLRs) in SE~\citep{kitchenham2009systematic}.

Studies were included based on the following criteria:

\begin{itemize}
    \item I1: Credibility: Publication in highly-ranked conferences or journals in the SE and NLP domains, as determined by a venue assessment.
    \item I2: Core Focus and Methodology: Must be an empirical study presenting concrete results related to toxicity detection or mitigation strategies in SE, specifically utilizing or developing LLM methods and relying on human-annotated ground truth.
    \item I3: Data Quality: For studies utilizing human-annotated data, inter-annotator agreement must be at least $0.6$ for Cohen's $\kappa$~\citep{cohen1960coefficient} or $0.8$ for Krippendorff's $\alpha$~\citep{krippendorff2018content}.
\end{itemize}

Studies were excluded based on the following criteria:

\begin{itemize}
    \item E1: Topic Mismatch: Purely conceptual works, or studies not directly focused on toxicity within SE.
    \item E2: Language Mismatch: Non-English language papers.
    \item E3: Duplicates: Identification as a duplicate entry across databases.
\end{itemize}

\begin{figure}[htp]
    \centering
    \includegraphics[width=13cm]{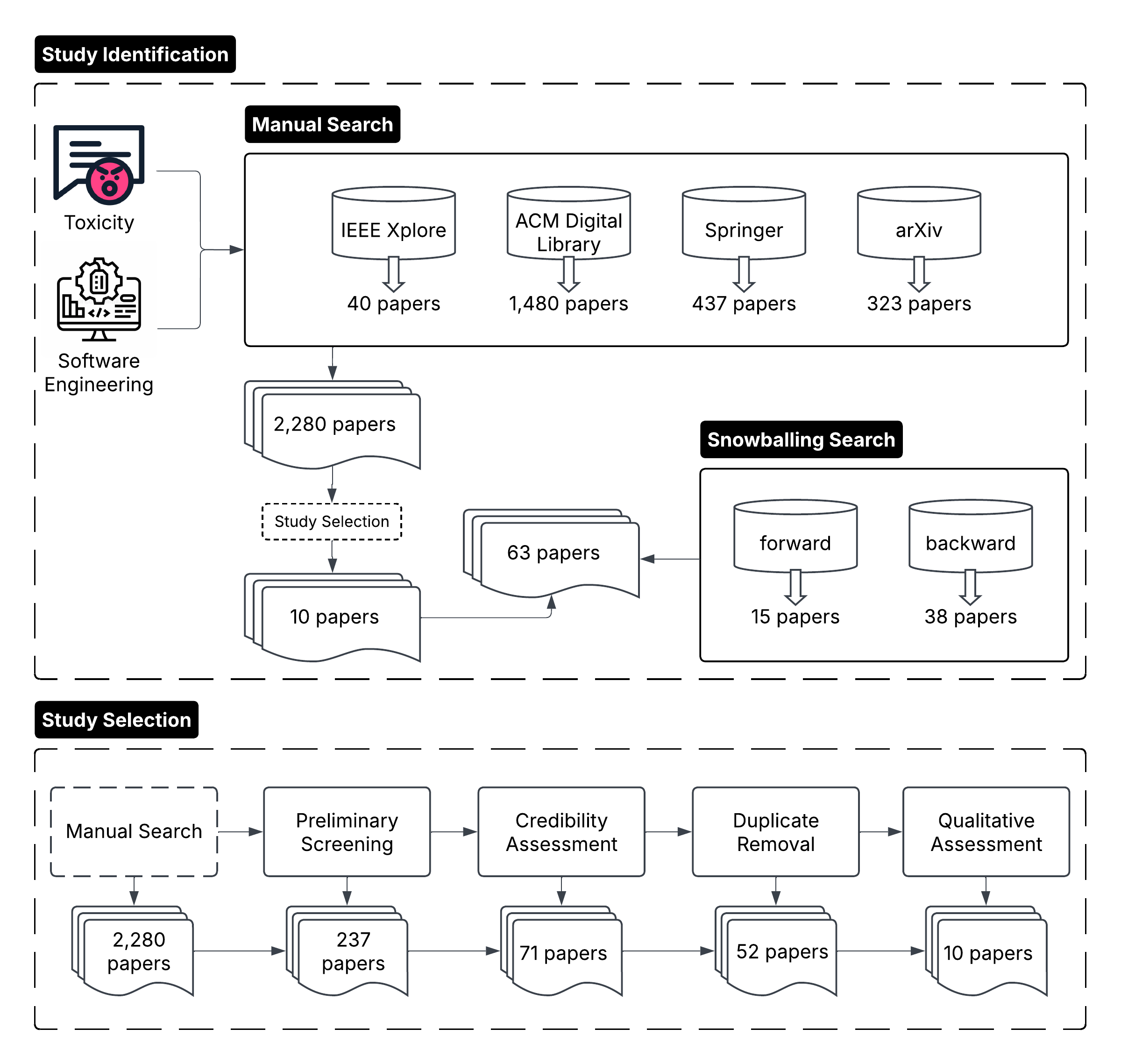}
    \caption{Study identification and selection process applied in this literature review}
    \label{fig:Study_Selection}
\end{figure}

This rigorous, four-step process is outlined in Table ~\ref{tab:study-selection}, resulting in a final corpus of 63 unique and highly relevant primary studies for this survey. The complete process is illustrated in Figure~\ref{fig:Study_Selection}.

\begin{table*}[!htbp]
\small
\caption{Queries to four digital libraries to identify relevant works}
\label{tab:library-queries}
\centering
\resizebox{\textwidth}{!}{%
\begin{tabular}{@{}p{2cm}p{9cm}p{4cm}p{1cm}@{}}
\toprule
\textbf{Library} & \textbf{Query String}                                                                           & \textbf{Filters}            & \textbf{Result} \\
\midrule
IEEE Xplore & (``All Metadata":toxic large language models)                                                   & Conferences; 2020-2024      & 40     \\
ACM         & {[}All: toxic large language models{]} AND {[}E-Publication Date: (01/01/2020 TO 12/31/2024){]} & Research Article            & 1480   \\
Springer    & toxic large language models)                                                                    & Conference paper; 2020-2024 & 437    \\
arXiv &
  order: -announced\_date\_first; size: 50; date\_range: from 2020-01-01 to 2024-12-31; classification: Computer Science (cs); include\_cross\_list: True; terms: AND all=toxic large language models &
  N/A &
  323 \\
\bottomrule
\end{tabular}%
}
\end{table*}

\begin{table*}[!htbp]
\small
\centering
\caption{Summary of the four-step study selection process}
\label{tab:study-selection}
\resizebox{\textwidth}{!}{%
\begin{tabular}{@{}l p{6cm} c c c@{}}
\toprule
\textbf{Step} & \textbf{Process Description} & \textbf{Initial Pool} & \textbf{Excluded} & \textbf{Resulting Pool} \\
\midrule
1 & Preliminary Screening: Title, abstract, and keyword screening (E1, E2) was performed to ensure basic relevance, language, and publication time. & 2,280 & 2,043 & 237 \\
\midrule
2 & Credibility Assessment: A full-text evaluation and publication venue check was performed against criterion I1, prioritizing highly-ranked venues. & 237 & 166 & 71 \\
\midrule
3 & Duplicate Removal: A check for unique entries was conducted against criterion E3. & 71 & 19 & 52 \\
\midrule
4 & Qualitative Assessment \& Snowballing: The 52 unique studies underwent a thorough qualitative assessment against criteria I2 and I3, focusing on LLM methods and data quality, resulting in 10 core studies, as detailed in Table \ref{tab:core-literature}. A subsequent snowballing approach was applied to this core literature. & 52 & 42 & 63 (after snowballing) \\
\bottomrule
\end{tabular}
}
\end{table*}

\begin{table*}[!htbp]
\centering
\caption{Overview of core literature on toxicity study in software engineering.}
\resizebox{\textwidth}{!}{%
\begin{tabular}{@{}V{1.5cm}V{6cm}V{2cm}V{2cm}V{6cm}V{2cm}V{2cm}@{}}
\toprule
    \textbf{Index \#} &
    \textbf{Paper Title} & 
    \textbf{Publication Venue} & 
    \textbf{Publication Year} & 
    \textbf{Covered Domains} & 
    \textbf{Detection} & 
    \textbf{Mitigation} \\ 
\midrule

1          & ToxiSpanSE: An explainable toxicity detection in code review comments                                                                                           & ESEM            &  ~\citeyear{sarker2023toxispanse}                             & SE datasets, Human annotator, Span-based detector, Domain-specific tailoring, LLM based model, Feature engineering & Yes       & No         \\
    \hline
2   & A benchmark study of the contemporary toxicity detectors on software engineering interactions                                                                   & APSEC           & ~\citeyear{sarker2020benchmark}                             & SE datasets, Non-SE datasets, Human annotator, Algorithmic annotator, Domain-specific tailoring                    & Yes       & No         \\
    \hline
3& Exploring ChatGPT for Toxicity Detection in GitHub                                                                                                              & ICSE            &  ~\citeyear{mishra2024exploring}                             & SE datasets, LLM based model, Domain-specific tailoring, Human annotator                                          & Yes       & No         \\
    \hline
4& Stress and burnout in open source: Toward finding, understanding, and mitigating unhealthy interactions                                                         & ICSE            &  ~\citeyear{raman2020stress}                             & SE datasets, Domain-specific tailoring, Human annotator, Feature engineering                                       & Yes       & No        \\
    \hline
5& An Empirical Evaluation of the Zero-shot, Few-shot, and Traditional Fine-tuning Based Pretrained Language Models for Sentiment Analysis in Software Engineering & IEEE Access     &  ~\citeyear{shafikuzzaman2024empirical}                             & SE datasets, Human annotator, Feature engineering, LLM based model, Domain-specific tailoring                      & Yes       & No         \\
    \hline
6 & Identification and mitigation of toxic communications among open source software developers                                                                     & ASE             &  ~\citeyear{sarker2022identification}                             & SE datasets, Human annotator, Data cleaning, Feature engineering, Domain-specific tailoring         & Yes       & No        \\
    \hline
7 & Towards offensive language detection and reduction in four software engineering communities            & EASE            &  ~\citeyear{cheriyan2021towards}                             & SE Datasets, Human annotator, Feature engineering, Domain-specific tailoring                                       & Yes       & Yes        \\
    \hline
8 & Realtoxicityprompts: Evaluating neural toxic degeneration in language models                                                                                    & arXiv preprint  &  ~\citeyear{gehman2020realtoxicityprompts}                             & Non-SE datasets, Algorithmic annotator, Data cleaning, LLM based model                                             & Yes       & No         \\
    \hline
9  & Exploring ChatGPT for identifying sexism in the communication of software developers               & PETRA           &  ~\citeyear{sultana2024exploring}                             & LLM based model, Prompt engineering                                                                              & Yes       & No         \\
    \hline
10  & DeMod: A Holistic Tool with Explainable Detection and Personalized Modification for Toxicity Censorship                                                         & arXiv preprint  &  ~\citeyear{li2024demod}                             & Non-SE Datasets, Human annotator, LLM based model, Domain-specific tailoring, Span-based detector                 & Yes       & Yes       \\
\hline
\label{tab:core-literature}
\end{tabular}%
}
\end{table*}

\begin{table*}[!htbp]
\large
\centering
\caption{Toxicity related datasets from SE community}
\resizebox{\textwidth}{!}{%
\begin{tabular}{@{}V{4cm}V{3cm}V{4cm}V{5cm}V{1cm}V{3cm}V{1cm}V{2cm}V{4cm}@{}}

\toprule
    \textbf{Dataset Name} &
    \textbf{Toxicity Task} & 
    \textbf{Data Source} & 
    \textbf{Source URL} & 
    \textbf{Size} & 
    \textbf{Annotation} & 
    \textbf{Paper} & 
    \textbf{Publication Venue} & 
    \textbf{Measurement} \\ 
\midrule
    ToxiCR Dataset & Toxicity Detection & FOSS project code reviews and Gitter chat messages & \url{https://github.com/WSU-SEAL/ToxiCR} & 20k & Manual labeling (2 of the authors) & ~\citeyear{sarker2022identification} & ASE & Accuracy, Precision, Recall, F-score, Cohen’s $\kappa$ \\
    \hline
    Benchmark Study Dataset & Toxicity Detection & FOSS project code reviews, Gitter chat messages, and Jigsaw test sample & \url{https://github.com/WSU-SEAL/toxicity-dataset/} & 14k & Manual labeling (2 of the authors) & ~\citeyear{sarker2020benchmark} & APSEC & Accuracy, Precision, Recall, F-score, Cohen’s $\kappa$ \\
    \hline
    ToxiSpanSE & Toxic Span Detection & ToxiCR Dataset & \url{https://github.com/WSU-SEAL/ToxiSpanSE} & 23k & Manual labeling (2 independent annotators) & ~\citeyear{sarker2023toxispanse} & ESEM & Krippendorff’s $\alpha$, F1 \\
    \hline
    GitHub Issues Dataset & Toxicity Detection & GitHub & \url{https://github.com/CMUSTRUDEL/toxicity-detector} & 2k & Perspective API & ~\citeyear{raman2020stress} & ICSE & F0.5 \\
    \hline
    Comment Dataset & Toxicity Detection & GitHub, Gitter, Slack, Stack Overflow & \url{https://archive.org/details/stackexchange} & 1,130k & Perspective API and Regex & ~\citeyear{cheriyan2021towards} & EASE & Precision \\
\bottomrule
\label{tab:toxicity-se-datasets}
\end{tabular}%
}
\end{table*}

\begin{table*}[!htbp]
\large
\centering
\caption{Toxicity related datasets from non-SE community}
\resizebox{\textwidth}{!}{%
\begin{tabular}{@{}V{4cm}V{4cm}V{3cm}V{5cm}V{1cm}V{3cm}V{1cm}V{2cm}V{4cm}@{}}
\toprule
    \textbf{Dataset Name} &
    \textbf{Toxicity Task} & 
    \textbf{Data Source} & 
    \textbf{Source URL} & 
    \textbf{Size} & 
    \textbf{Annotation} & 
    \textbf{Paper} & 
    \textbf{Publication Venue} & 
    \textbf{Measurement} \\ 
\midrule
    Real Toxicity Prompts & Toxicity Detection & Reddit & \url{https://github.com/allenai/real-toxicity-prompts} & 22k & Perspective API & ~\citeyear{gehman2020realtoxicityprompts} & EMNLP & Expected Maximum Toxicity and Toxicity Probability \\
    \hline
    Writing Prompts & Toxicity Mitigation & Reddit & \url{https://www.kaggle.com/datasets/ratthachat/writing-prompts} & 300k & Toxicity detection API & ~\citeyear{xu2022leashing} & AAAI & Toxicity Reduction \\
    \hline
    Offensive Language Identification Dataset (OLID) & Offensive Language Detection & Twitter & \url{https://sites.google.com/site/offensevalsharedtask/olid} & 14.2k & 3 annotators per rater group per comment & ~\citeyear{zampieri2019predicting} & NAACL & Per-class Precision, Recall, F1, Weighted Average \\
    \hline
    Hate Speech Dataset & Hate Speech, Offensive Language Detection & Twitter & \url{https://github.com/BenchengW/Offensive-Language-Detection-DL} & 25k & 3 to 9 annotators per post & ~\citeyear{touahri2020offensive} & AI2SD & Accuracy \\
\bottomrule
\end{tabular}%
}
\label{tab:toxicity-non-se-datasets}
\end{table*}


\section{Methodology}\label{methodology}
This section outlines common methodologies for addressing toxicity in SE. We first discuss annotation techniques involving human and algorithmic annotators. Next, we describe pre-processing methods, including data cleaning, feature engineering and sampling techniques. Finally, we discussed various models from multiple perspectives, including general versus domain-specific, model architecture, and task types.
\subsection{Annotation}
Accurate toxicity annotation is essential for building high-quality datasets, as the subjective nature of toxicity often requires nuanced human interpretation. This section explores various annotation methods, including human annotation and algorithmic approaches, while evaluating their respective strengths and limitations.

\subsubsection{Human annotator}
The involvement of human annotators is indispensable for capturing context-specific nuances in toxic content. To enhance annotation quality, researchers employ several strategies. One widely adopted approach is \textbf{the use of multiple annotators} with diverse backgrounds, which helps mitigate individual biases and improve perspective diversity. For instance,~\citet{sarker2023toxispanse} paired annotators of different genders to label samples, increasing the likelihood of balanced viewpoint representation.

To ensure consistent labeling, annotators often follow \textbf{a well-defined rubric} that outlines toxicity criteria. These rubrics can be pre-established or developed specifically for a given research context. For example, in an effort to tailor annotation guidelines to the SE domain,~\citet{sarker2020benchmark} had a group of annotators analyze 1000 SE texts and empirically develop a rubric for identifying toxicity. This process not only clarified labeling standards but also reduced subjectivity by aligning annotators with detailed, research-specific rules. The resulting rubric was later reused in subsequent work~\citep{sarker2022identification}, demonstrating its reliability and applicability across studies.

Beyond following structured guidelines, researchers also assess annotation quality through \textbf{inter-annotator agreement} metrics such as Cohen’s $\kappa$ or Krippendorff’s $\alpha$. The choice of metric depends on the complexity of the task and the specific research needs. For instance,~\citet{sarker2023toxispanse} utilized Krippendorff’s $\alpha$, which accounts for multiple annotators, missing values, and partial agreements—making it particularly suitable for nuanced tasks like toxicity annotation. By quantifying consistency between annotators, these metrics help validate dataset reliability. When disagreements arise, \textbf{conflict resolution mechanisms} are employed to establish a reliable ground truth. Typically, a third annotator or a consensus process determines the final label, ensuring that conflicting perspectives are reconciled. 

For practical implementation in annotation teams, particularly within the SE domain, establishing clear protocols for \textbf{annotator training}, target inter-annotator agreement (IAA), and conflict resolution is crucial. Training protocols often involve initial calibration sessions where annotators discuss guidelines and ambiguities, ensuring a shared understanding of criteria. While specific benchmarks for IAA can vary, aiming for high agreement is essential for data reliability. For instance, Cohen's $\kappa$ is typically considered to indicate substantial agreement above 0.7, and studies like~\citet{sarker2023toxispanse} have reported agreement as high as 0.96. When disagreements arise, an adjudication workflow, such as facilitated discussions or involvement of a senior annotator, is necessary to reach consensus and establish a robust ground truth for conflicting labels.

Additionally, \textbf{multi-label annotation} offers a more nuanced alternative to traditional binary classification, which often oversimplifies the complexity of toxic language. Instead of strictly categorizing content as offensive or non-offensive, multi-label annotation captures contextual dependencies and annotators' varying perspectives. ~\citet{arhin2021ground} introduced an approach incorporating three label types—normal, offensive, and undecided—along with three labeling schemes (strict, relaxed, and inferred group labels). This method enables a more comprehensive and context-aware annotation process, ultimately leading to more robust toxicity detection models.

\subsubsection{Algorithmic annotator}
To reduce the costs and time associated with manual annotation, algorithmic tools are increasingly employed in toxicity research. These tools enable efficient labeling of large-scale corpora, providing automated assessments that assist researchers in filtering and prioritizing data for analysis. Among these, several \textbf{commercial APIs} and \textbf{open-source libraries} offer automated assessment capabilities.

One widely used tool in this domain is the Google Perspective API~\citep{conversationAI}, which assigns toxicity scores to text based on linguistic attributes such as ``toxicity", ``insult,`` and ``identity attack". Developed by Google Jigsaw, this API leverages machine learning models trained on extensive datasets to identify harmful or offensive language patterns in online communication. Beyond Perspective API, other notable commercial services include IBM Watson's Tone Analyzer, which offers sentiment and tone analysis for various communication styles~\citep{tone2018}, and Microsoft's Azure AI Content Safety (formerly Content Moderator), designed to detect objectionable content across text and images~\citep{azure2025}. Researchers also widely utilize open-source libraries such as Detoxify, which provides flexible and customizable solutions for toxicity classification~\citep{detoxify2021}.

The Perspective API is particularly useful for pre-labeling datasets, allowing researchers to streamline manual annotation efforts while gaining a granular understanding of toxicity intensity. Its scoring system facilitates the evaluation of language models and informs the development of mitigation strategies. Given its effectiveness, studies~\citep{cheriyan2021towards, gehman2020realtoxicityprompts, raman2020stress} frequently use Perspective API as a benchmarking tool for validating toxicity detection methods. Additionally, it is often integrated into hybrid annotation pipelines that combine algorithmic and human inputs, thereby improving data quality and model robustness.

However, despite its widespread adoption, the Perspective API has notable limitations and potential biases that impact its reliability and fairness. Studies have highlighted systematic biases in its scoring, particularly against marginalized groups. For example,~\citet{sap2019risk} demonstrated that the API disproportionately flags African American Vernacular English (AAVE) as more toxic than standard English, underscoring the risks of algorithmic discrimination in automated moderation tools. Furthermore, the API’s lack of transparency in its training data and model updates presents challenges for reproducibility and fairness auditing.~\citet{bender2021dangers} emphasize that black-box models like Perspective API hinder researchers' ability to critically assess how biases emerge, making it difficult to address or mitigate them. Without open access to the underlying data, researchers must rely on empirical testing to infer weaknesses, which limits their ability to propose targeted improvements.

In addition to machine learning-based approaches, researchers also employ simpler rule-based methods such as \textbf{regular expression (regex) matching} to detect explicit toxic content. Regex allows for the definition of patterns that capture specific linguistic constructs, including slurs, profanities, or abusive language, enabling rapid identification of harmful text. For instance, regex can be used to flag messages containing offensive words, including obfuscated spellings such as ``f**k." The primary advantage of regex-based methods lies in their speed and cost-effectiveness, as they require minimal computational resources compared to more sophisticated machine learning models or APIs. However, regex-based approaches have significant limitations: they struggle to interpret context, irony, or subtle variations in language, often leading to high false positive or negative rates.~\citet{nozza2023state} discussed the complexities introduced by various obfuscation methods, such as character substitutions, which complicate the detection process and require regex patterns to account for numerous variations. For example, crafting regex patterns to identify variations like ``f**k" requires anticipating and encoding potential character substitutions. Despite these drawbacks, regex remains a valuable tool for filtering explicit toxicity, particularly when used in conjunction with more advanced annotation techniques.

\subsection{Pre-processing}
Pre-processing plays a pivotal role in reducing noise and standardizing inputs. A recent study~\citep{siino2024text} investigates the impact of popular pre-processing methods on transformer models. The findings indicate that, in some cases, with an appropriate pre-processing strategy, even a simple Naïve Bayes classifier can outperform the best-performing transformer by 2\% in accuracy. Other studies provide direct evidence that certain pre-processing techniques enhance the effectiveness of hate and offensive speech detection~\citep{glazkova2023comparison} as well as toxicity detection~\citep{mohammad2018preprocessing}.

\subsubsection{Data cleaning}
Effective data cleaning is essential for pre-processing text in toxicity detection tasks, as extraneous and misleading content can interfere with model performance. One common step is \textbf{URL removal}, as URLs often introduce external references that may skew predictions. URLs are typically detected using regex patterns that match common formats such as ``http://" or domain names (e.g., ``www.example.com"). Removing these links prevents models from relying on unrelated external content, ensuring they focus solely on textual features. This technique has been widely applied in prior studies~\citep{jemai2021sentiment, athar2021sentimental} to reduce biases. However, its overall effectiveness remains a topic of discussion. Studies by~\citet{jianqiang2017comparison}, for instance, observe that URL removal often has a negligible impact on sentiment classification accuracy, with fluctuations in accuracy and F1-measure generally limited to +/-0.5\% and, in many cases, no change at all. This suggests that it is often outperformed by more advanced pre-processing techniques.

Another important step is \textbf{contraction expansion}, where informal contractions like ``can't" or ``won't" are expanded into their full forms (``cannot" and ``will not"). This transformation standardizes text, making it easier for models to parse words correctly and recognize them as distinct entities. Studies have shown that contraction expansion improves accuracy and F1-scores in sentiment classification, demonstrating its relevance in text pre-processing pipelines. For instance,~\citet{jianqiang2017comparison} report accuracy gains of up to 6.08\% and F1-score improvements of up to 6.85\% on certain datasets and classifiers after expanding acronyms. However, its effectiveness can vary across domains, as expanding contractions may introduce unnecessary verbosity or alter the tone of informal text, potentially impacting sentiment analysis in conversational contexts~\citep{aadil2023exploring}.

\textbf{Symbol removal} is another pre-processing step that enhances text clarity. Symbols such as ``@", ``\#", or excessive punctuation (e.g., ``!!!") are common in social media text but do not necessarily contribute meaningful information for toxicity detection. Their presence can introduce noise and cause models to overfit to irrelevant patterns. Removing unnecessary symbols~\citep{anbukkarasi2020analyzing, tan2023survey} helps streamline the data and ensures that predictions are based on meaningful language rather than stylistic artifacts. Similarly, repetition elimination addresses exaggerated expressions commonly found in informal text, where characters (e.g., ``loooove" → ``love") or words (e.g., ``really really good") are repeated. If not addressed, models may treat these variations as distinct features, leading to unnecessary complexity. By simplifying repeated elements~\citep{aziz2020twitter}, we maintain the intended meaning while improving model consistency. 

However, it's important to note that while symbol removal can reduce noise, it may also strip away context that is crucial for understanding the tone or intent behind a message. For instance, emoticons often convey contextual meaning, which can be valuable for sentiment analysis~\citep{aadil2023exploring}. Thus, indiscriminate removal of symbols may risk losing important nuances in informal text, particularly on platforms where symbols play a significant role in communication. This nuanced impact on model performance is evident from studies like~\citet{jianqiang2017comparison}, which reports that while some models gained up to 12.52\% in F1-score after reverting repetitions, others experienced significant losses, such as a 12.11\% accuracy decrease for Naive Bayes on certain datasets.

For \textbf{domain-specific datasets}, particularly those containing programming-related text, additional pre-processing steps are necessary to distinguish between toxic language and structural code components. One such technique is \textbf{identifier splitting}, which breaks down compound variable names into separate tokens. Many programming identifiers, such as ``isToxic", use camelCase or snake\_case formatting, making them difficult for models trained on natural language text to interpret. Splitting these identifiers into distinct words (``is" and ``toxic") allows models to focus on meaningful linguistic elements rather than code syntax, enhancing their ability to detect toxicity. This approach has shown positive impacts on model performance; in the context of code similarity detection,~\citet{karnalim2020preprocessing} report F-score increases of around 14\% and recall improvements of around 35\% on Python datasets after applying identifier removal.

Another pre-processing method specific to code-related datasets is \textbf{programming keyword removal}. Programming languages contain syntax elements such as keywords (``if", ``else", ``return"), operators (``+", ``=", ``\&\&"), and delimiters (``\{", ``\}", ``;"), which are crucial for code execution but generally irrelevant for toxicity detection. By removing these elements, models can concentrate on meaningful text components, such as comments or string literals, where toxicity is more likely to be present. This approach has been validated in studies like~\citet{hacohen2020influence}, which identified the removal of HTML objects and syntax elements as a key step in effective pre-processing. Filtering out programming-specific tokens helps models assess toxicity more accurately by preventing structural code artifacts from influencing the analysis.

To guide practitioners in applying these pre-processing techniques, we propose a sequenced workflow, particularly for toxicity detection in software engineering contexts. A general pipeline would typically involve: 1) removing URLs, 2) expanding contractions, 3) eliminating repetitions, and 4) performing symbol removal. For domain-specific datasets, such as those containing programming-related text, additional steps would then be applied: 5) splitting identifiers, and 6) filtering out programming keywords. The precise order and inclusion of steps may vary based on dataset characteristics and model requirements, but this general sequence provides a robust starting point for effective data cleaning.

\subsubsection{Feature engineering}
Feature engineering plays a crucial role in improving model performance by incorporating customized attributes that enhance the detection of toxic content. One key technique is \textbf{adversarial pattern identification}, focusing on detecting and mitigating inputs designed to bypass toxicity filters, such as obfuscated slurs or profanities using patterns like \texttt{f[\textbackslash\textbackslash*@\_\textbackslash\textbackslash-]+ck} for ``fuck" or \texttt{s\#+it} for ``shit" that account for character substitutions. For instance,~\citet{hartvigsen2022toxigen} employ adversarial pattern generation to identify implicit toxicity, enhancing classifier robustness against evasive language. However, this approach risks inadvertently perpetuating bias by focusing on specific, often marginalized, groups targeted by toxic language, potentially amplifying the very patterns it aims to detect rather than addressing root causes. A study~\citep{sap2019risk} further demonstrates that toxicity detection models trained on existing datasets often exhibit significant performance disparities across demographic groups, with higher false positive rates for language associated with minority identities. This indicates that a narrow focus on adversarial pattern identification, without considering broader social context and potential biases, can exacerbate existing inequities in toxicity detection.

\textbf{Profane word counting} is another widely used approach, which calculates the density of harmful or offensive words in a comment, typically as the ratio of identified profane words to the total word count. This numerical feature serves as a quantifiable indicator of potential toxicity for classifier training. For example, research by~\citet{bhat2021say} outlines a taxonomy for workplace toxicity and highlight profanity as a key indicator of offensive language. Similarly,~\citet{d2019towards} include profane word counting in their pre-processing pipeline to enhance classification accuracy.

\textbf{Embedding features} leverage word or sentence embeddings, such as Word2Vec, GloVe~\citep{zhang2019hate}, and BERT~\citep{caselli2020hatebert}, to represent text in a high-dimensional continuous space. These embeddings capture semantic relationships between words, allowing models to better understand nuanced language patterns associated with toxicity. By transforming text into a structured numerical format, embedding features enhance the model's ability to detect implicit and context-dependent toxic language. Embedding features, while powerful for capturing semantic relationships, risk amplifying societal biases present in their training data. Studies like~\citet{blodgett2020language} reveal that such biases can lead to disproportionate toxicity flagging of language associated with marginalized groups, necessitating careful mitigation strategies.

Finally, \textbf{contextual features} are crucial for analyzing user interactions and conversational history to disambiguate toxicity. Isolated toxic statements may appear ambiguous without considering the surrounding discussion. For example,~\citet{pavlopoulos2020toxicity} investigate the importance of conversational context in toxicity classification, demonstrating that models incorporating thread structure and user history outperform those relying solely on individual comments. By integrating contextual features, models can differentiate between harmful intent and benign language, improving overall classification accuracy. However, processing long conversational histories introduces a significant computational burden, hindering real-time applications due to the memory and processing demands of capturing long-range dependencies. Research into transformer efficiency~\citep{tay2022efficient} highlights the ongoing efforts to address this challenge.

Beyond the generation of effective features, careful feature selection and dimensionality reduction are critical steps to prevent overfitting, reduce computational complexity, and enhance model interpretability. Techniques such as mutual information~\citep{battiti1994using} and chi-square tests~\citep{yang1997comparative} can be employed to evaluate the statistical dependence between features and the target variable, thereby identifying the most predictive attributes. Additionally, using L1-regularized models (e.g., Lasso regression~\citep{tibshirani1996regression}) inherently performs feature selection by shrinking the coefficients of less important features to zero. Implementing these strategies guides practitioners in curating a lean yet powerful feature set, optimizing model performance and generalization capabilities.

\subsubsection{Sampling techniques}
To ensure balanced representation of toxic and non-toxic samples, studies often employ \textbf{stratified sampling} based on external toxicity scores, such as those from Google’s Perspective API~\citep{miller2022did, sarker2020benchmark}. Given the rarity of toxic content, practical guidelines involve stratifying samples into specific toxicity score ranges; for instance,~\citet{sarker2020benchmark} used 0.1-interval groups for scores below a 0.5 threshold to ensure diverse representation across the non-toxic spectrum. This approach balances the distribution of toxicity likelihoods across subsets, exposing the model to diverse samples and preventing overfitting to a single class. However, while stratified sampling prioritizes representativeness, it may not always select the most informative samples for model training. To tackle data imbalance, other over-sampling techniques like SMOTE~\citep{chawla2002smote} can be utilized.

As an alternative, \textbf{active learning} strategically selects the most valuable data points for labeling, aiming to maximize model learning with minimal labeled data. For instance,~\citet{figueroa2012active} demonstrated that active learning can achieve 90\% accuracy with approximately 33\% fewer instances than random sampling on certain datasets. Beyond just efficiency, active learning also offers an advantage in addressing data imbalance, as it focuses on efficiently acquiring informative samples from the minority class. Consequently, active learning could serve as an effective alternative or complementary technique to stratified sampling, particularly when labeling costs are a significant consideration.

\subsection{Detection \& Mitigation}
The detection and mitigation of toxicity in SE communication have evolved dramatically. It has moved from simple, keyword-based methods to sophisticated, context-sensitive models. Current strategies balance the broad domain coverage of general-purpose models with the higher precision of SE-tuned, domain-specific ones. Given the unique terminology and nuanced patterns of SE text, recent work underscores the need for SE-specific pre-processing and fine-tuning. In the sections below, we first compare general-purpose and SE-specific approaches, then examine different architectures and task-level methods for toxicity management in SE.
\subsubsection{General vs. Domain-Specific Toxicity}
Large pre-trained language models, such as GPT-3, GPT-4, and Gemini, demonstrate notable flexibility in toxicity detection tasks due to their zero-shot and few-shot capabilities.~\citet{mishra2024exploring} showcased how ChatGPT effectively identified toxic comments on GitHub without requiring domain-specific fine-tuning, leveraging its training on diverse datasets. Similarly,~\citet{sultana2024exploring} demonstrated ChatGPT’s success in detecting gender biases and stereotypes, such as maternal insults and subtle stereotyping, in SE communications. Generative pre-training enables these models to adapt to a range of tasks with minimal input. However, performance evaluations reveal limitations:~\citet{shafikuzzaman2024empirical} showed that while GPT-4o performs reasonably well in smaller SE datasets, it falls behind fine-tuned models like seBERT when applied to larger datasets such as GIT and ARC. This highlights a critical trade-off between flexibility and performance when domain-specific nuances are involved.

\begin{figure}[htp]
    \centering
    \includegraphics[width=8cm]{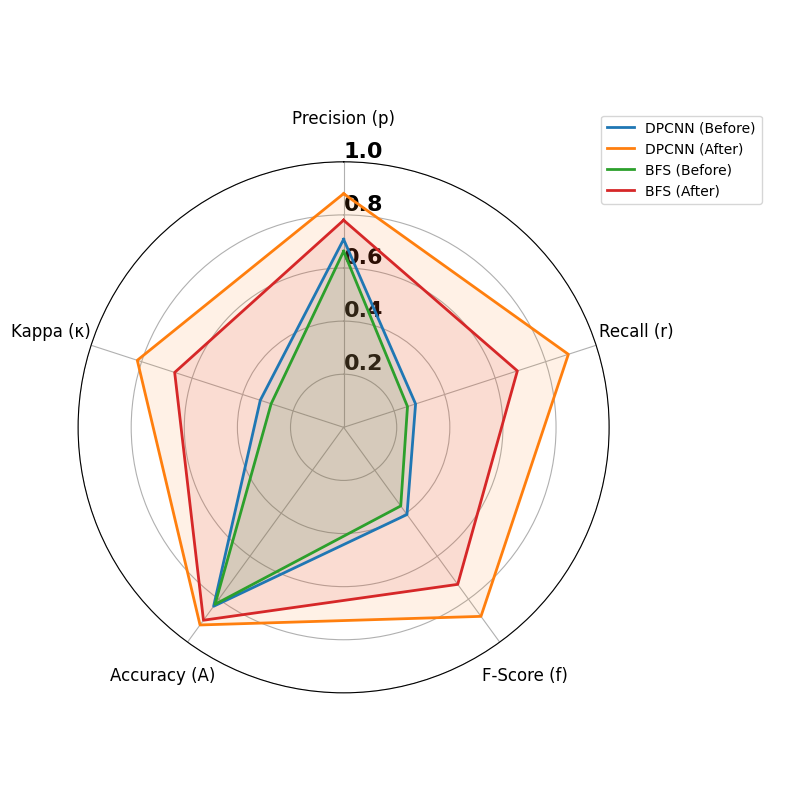}
    \caption{Code review dataset (comparing methods before \& after retrained on SE Dataset)}
    \label{fig:Code_Review_Dataset}
\end{figure}

\begin{figure}[htp]
    \centering
    \includegraphics[width=8cm]{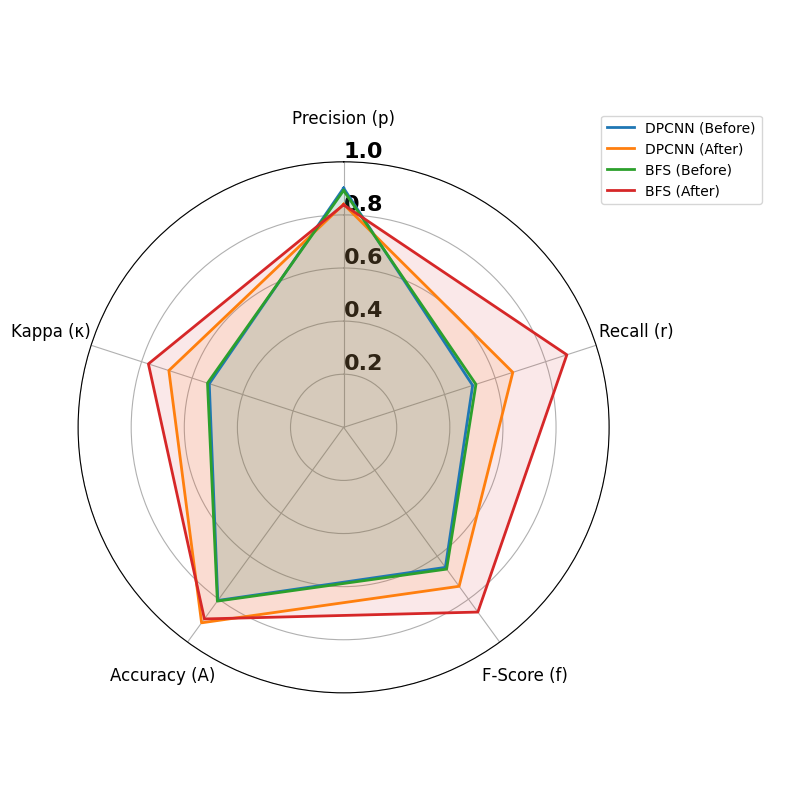}
    \caption{Gitter Ethereum dataset (comparing methods before \& after retrained on SE dataset)}
    \label{fig:Gitter_Ethereum_Dataset}
\end{figure}

Focusing on the specific requirements of SE discussions, domain-specific models fine-tuned on SE datasets consistently outperform general-purpose models in both accuracy and contextual relevance.~\citet{sarker2022built} emphasized the importance of pre-processing SE-specific language, noting that general-purpose models often misclassify technical terms such as ``kill" or ``junk" as toxic. The other benchmark study demonstrated how retraining in SE-specific data significantly improves performance. For example,~\citet{sarker2020benchmark}'s experiments evalutated tools like DPCNN and BFS after retraining on SE datasets such as Code Review and Gitter Ethereum revealed considerable improvements in key metrics such as precision, recall, and F score. The accompanying Figure~\ref{fig:Code_Review_Dataset} and Figure~\ref{fig:Gitter_Ethereum_Dataset} highlight these findings, with F scores for DPCNN and BFS reaching 0.406 and 0.366, respectively, on Code Review, and 0.652 and 0.660 on Gitter Ethereum. These metrics underscore the value of retraining models to handle SE-specific language more effectively.

In addition,~\citet{sarker2023toxispanse} introduced a span-based toxicity detector that identifies specific toxic phrases in the SE text, providing improved interpretability for moderators. Beyond binary classification,~\citet{cheriyan2021towards} analyzed toxicity across platforms like GitHub and Stack Overflow, advocating for SE-specific context-aware solutions to reduce conflicts and enhance community dynamics. Recent benchmarks also highlight seBERT’s macro F1-scores exceeding 90\% on datasets like GIT, reflecting its ability to capture domain-specific nuances with unparalleled precision~\citep{shafikuzzaman2024empirical}. These examples emphasize that models tailored to the SE domain are essential for achieving superior performance in detecting and mitigating toxicity.


\begin{table}[htbp]
\caption{Cloud LLM APIs and corresponding operational costs}
\label{tab:llm-price}
\centering
\begin{tabular}{l l r r r}
\toprule
\textbf{Model} & \textbf{Provider} & \textbf{Input (\$/M tokens)} & \textbf{Output (\$/M tokens)} & \textbf{Latency (p99, s)} \\
\midrule
GPT-4.1           & OpenAI   & 2.00 & 8.00 & 2.66 \\
Gemini 1.5 Flash  & Google   & 0.60 & 2.40 & 0.92 \\
\bottomrule
\end{tabular}
\end{table}

Table~\ref{tab:llm-price} summarizes the prices and latencies of the commercial LLM APIs used in our experiments in Section \ref{ablation}, reflecting the conditions at the time the experiments were conducted~\citep{openai2024gpt41, Google2025VertexAIPricing}. For the latest and most accurate information regarding model availability and pricing, please refer to the providers’ official websites, as these models are frequently updated and iterated. Because these services are pay-as-you-go, teams can simply pick the SKU that matches their quality target and budget. Take GPT-4.1 as an example: suppose we need to moderate 60 comments every hour, each averaging 50 English words. OpenAI’s tokenizer turns English prose into roughly one token for every 0.75 word, so a 50-word comment is about 67 tokens. Multiplying by 60 comments yields around 4,000 input tokens per hour. If we budget a short, one-token label plus a few JSON characters—say 5 output tokens per comment—that adds another 300 output tokens. The total comes to about \$0.01 per hour.

A second route is to deploy the model to cloud service providers like Amazon Web Service. An inexpensive yet GPU-equipped choice is g4dn.xlarge (1 $\times$ NVIDIA T4, 16 GB)~\citep{AWS2025G4dnXlarge}. The on-demand price in us-east-1 is \$0.526 per hour. Adding a 30 GB gp3 EBS volume for the OS and model weights costs about \$2.40 per month at the published \$0.08 GB-month rate. Network pricing is favourable: inbound traffic is free and the first 100 GB of data that leaves AWS each month is also free; only excess egress is billed (starting near \$0.09 GB). With seBERT’s $\sim$1 ms inference time per 128-token comment, one g4dn.xlarge can serve roughly one million requests in 20 minutes, so an eight-hour moderation shift adds about \$4 of compute.

The advantages and disadvantages of each option are presented in Table~\ref{tab:llm_deploy_choices}.

\begin{table*}[!htbp]
\small
\centering
\caption{Comparison of LLM deployment choices}
\label{tab:llm_deploy_choices}
\resizebox{\textwidth}{!}{%
\begin{tabular}{@{}p{4cm}p{7cm}p{7cm}@{}}
\toprule
\textbf{Deployment Choice} & \textbf{Advantages} & \textbf{Disadvantages} \\
\midrule
Commercial LLM API (e.g., GPT-4.1)
& Zero ops, instant scalability. Pay-as-you-go pricing (as low as \$0.01/h for the 60-comment example). No GPU provisioning required.
& Latency depends on network conditions. Costs grow linearly with usage (per-token charges). Data leaves your VPC. \\

Cloud Self-Host (EC2 g4dn.xlarge)
& Fixed hourly rate (\$0.526/h) with full GPU control. Domain-tuned models (e.g., seBERT) may achieve higher F1 at similar runtime cost. 100 GB outbound traffic/month is free.
& Charges apply even when idle (unless instance stopped). Requires manual patching, monitoring, and scaling. Additional EBS storage at \$0.08/GB-mo. \\

On-Premise Hardware
& Data never leaves local network. No recurring cloud fees after purchase. Sub-millisecond latency possible with PCIe GPUs.
& High upfront cost. Depreciation, power, and cooling overhead. Full responsibility for redundancy, security, and capacity planning. \\
\bottomrule
\end{tabular}%
}
\end{table*}

\subsubsection{Architectural difference}
Traditional models offer a straightforward approach to toxicity detection in SE discussions but often fail to account for contextual nuances. Early SE toxicity detection often relied on lexicon-based approaches or classical machine learning algorithms like SVMs. Lexicon-based approaches use predefined dictionaries of words or phrases associated with specific sentiments or toxicity levels. For example, terms like ``kill" or ``junk" might be flagged as toxic based on such a dictionary, without considering their contextual meaning~\citep{jongeling2017negative}. These methods, while easy to implement and understand, often misinterpret technical jargon or colloquial expressions common in SE communication, leading to a high rate of false positives~\citep{sarker2020benchmark}.~\citet{raman2020stress} demonstrated the use of an SVM tailored for GitHub issue comments, revealing its potential for identifying toxicity trends across communities, but its performance was limited by its reliance on simple feature sets. Moreover, empirical evaluations have shown that while lexicon-based models can achieve acceptable precision on datasets like Code Review and Gitter Ethereum, their recall remains poor when handling nuanced or domain-specific expressions~\citep{sarker2022identification, sarker2020benchmark}.

In addition to traditional models like lexicon, some attention-based methods have also been applied to the field of toxicity detection. These neural models explicitly learn to weigh tokens or sentences by importance, typically adding word- or sentence-level attention atop lightweight encoders (e.g., GRU or CNN–RNN hybrids) to spotlight toxic cues. \citet{pavlopoulos2017deeper} showed that RNN based on word vectors outperforms logistic regression or MLP classifiers with character or word n-grams in review. At the same time, this method also brings another benefit: suspicious words can be highlighted for free without including labeled words in the training data. Subsequent studies compared contextual vs.\ self-attention and confirmed the benefit of contextual attention for implicit abuse~\citep{chakrabarty2019pay}. Recently, an attention-enhanced CNN–LSTM achieved an accuracy of 0.76 on the shared task on Abusive Comment Detection in Tamil - ACL 2022, outperforming deep learning models like CNN, LSTM, Bi-LSTM ~\citep{first2024attention}.

LLM-based models capitalize on generative pre-training to develop richer contextual understanding. These methods utilize transformer-based architectures (BERT, RoBERTa, GPT-3, GPT-4), trained on large-scale, generative pre-training objectives. For example, generative pre-training involves training models to predict masked words or the next token in a sequence, enabling them to capture rich semantic relationships in text~\citep{kenton2019bert, tan2022roberta}. ToxiSpanSE~\citep{sarker2023toxispanse}, a fine-tuned RoBERTa-based model, achieved an impressive F1-score of 0.88 in detecting toxic spans in SE datasets, significantly outperforming earlier lexicon-based methods. Similarly,~\citet{shafikuzzaman2024empirical} evaluated GPT-3.5 and GPT-4, demonstrating their strong performance in zero-shot and few-shot settings on SE datasets like ARC and GIT. Another innovative example is the DeMod tool~\citep{li2024demod}, which incorporates explainable AI capabilities to enhance both detection and mitigation of toxicity. By leveraging ChatGPT's generative pre-training, DeMod offers fine-grained explanations and personalized modification suggestions to address toxic content.

In order to compare the commonly used transformer-based models and their phenotypes for toxicity detection, we conducted an experiment. Five Transformer variants are fine tuned under a uniform recipe (1 epoch, batch size 32, learning rate 2e-5, max sequence length 512, using an 80/20 split for the training and testing datasets)~\citep{devlin2019bert,liu2019roberta,trautsch2022predicting}. The dataset on software engineering code reviews is shared by \cite{sarker2023automated}. Table~\ref{tab:berts} summarizes each model’s architecture and classification performance. Among the base‐size Transformers, seBERT—built on the BERT-base backbone with SE-specific tokenization—achieves the highest accuracy (0.9463) and F1 score (0.9461). BERT-base follows closely (0.9456 accuracy, 0.9450 F1), while RoBERTa-base trails slightly (0.9420, 0.9414). Scaling up to the large variants under a single‐epoch, default‐hyperparameter regime yields no clear gains: BERT-large simply matches BERT-base (0.9456, 0.9450), and RoBERTa-large actually performs worse (0.9395, 0.9385). These results indicate that for mid-sized SE toxicity datasets, base-size models strike a better balance of capacity and stability. Larger models—with doubled depth and hidden dimensions—tend to underfit or train unstably when constrained to one epoch; fully leveraging their extra capacity therefore requires stronger optimization momentum, more training data, and additional epochs to achieve further performance improvements.

\begin{table*}[!htbp]
\small
\centering
\caption{Architecture and performance of BERT-family models}
\label{tab:berts}
\resizebox{\textwidth}{!}{%
  \begin{tabular}{@{}llllllll@{}}
    \toprule
    \textbf{Model} & \textbf{Parameters} & \textbf{Layers} & \textbf{Hidden Size} & \textbf{Heads} & \textbf{Embedding Size} & \textbf{Accuracy} & \textbf{F1} \\
    \midrule
    BERT-base       & 110 M  & 12  & 768  & 12  & 768  & 0.9456 & 0.9450 \\
    BERT-large      & 340 M  & 24  & 1024 & 16  & 1024 & 0.9456 & 0.9450 \\
    RoBERTa-base    & 125 M  & 12  & 768  & 12  & 768  & 0.9420 & 0.9414 \\
    RoBERTa-large   & 355 M  & 24  & 1024 & 16  & 1024 & 0.9395 & 0.9385 \\
    seBERT          & 110 M  & 12  & 768  & 12  & 768  & 0.9463 & 0.9461 \\
    \bottomrule
  \end{tabular}%
}
\end{table*}

Despite their impressive F1 scores, transformer-based toxicity detectors inherit two well-documented weaknesses. First, they are highly sensitive to adversarial perturbations that confuse malicious tokens through character-level perturbations. The detection recall rate of BERT's detector will be reduced by more than 50\% in some cases~\citep{kurita2019towards}, and recent benchmarks such as TaeBench record attack-success rates above 77\% against both open-source and commercial APIs~\citep{zhu2024taebench}.~\citet{nozza2023state} also showed that the various ``fancy writing" of swear words/abusive words is enough to significantly reduce the recall rate of mainstream transformer-based detectors. Second, detectors’ performance degrades sharply under domain shift. Empirical studies show that transformer models fine-tuned on the Jigsaw/Wikipedia Toxic Comment corpus lose 40–53\% of their F1 when applied to software-engineering data such as GitHub code review or Gitter messages~\citep{sarker2020benchmark}. The other complementary cross-platform analysis finds that fine-tuned language models keep only around 50\% of their in-domain performance under cross-domain testing~\citep{schouten2023cross}.

In response to these shortcomings, an open source tool PROF was developed, which is used to automatically generate a variety of ``fancy writing methods" and inject them into the training set, significantly improving the robustness of the model under such disturbances in a confrontational data enhancement manner~\citep{nozza2023state}.

\subsubsection{Task type}
Most toxicity detection approaches focus on a binary classification task, labeling each comment as toxic or non‑toxic. For instance, the five datasets from Table~\ref{tab:toxicity-se-datasets} are all examples of binary-classified datasets from the SE community. Models built using these datasets typically yield an overall yes/no decision without pinpointing which parts of the text are problematic. 

In contrast, span-level detection identify the exact tokens that carry toxic meaning. Span-based tools such as ToxiSpanSE highlight the problematic words inside code review comments, giving moderators an immediate rationale for the model’s decision~\citep{sarker2023toxispanse}. Table~\ref{tab:spannospan} summarizes the detection performance of ToxiCR and ToxiSpanSE. The data is selected from their fine-tuned model based on BERT, and they use the same Code Review dataset. From the data, ToxiCR slightly outperforms in overall classification. However, ToxiSpanSE provides the added benefit of fine-grained localization of toxic tokens. In practice, the two can be combined: first apply ToxiCR for high-throughput filtering, then invoke ToxiSpanSE on flagged comments to obtain precise, explainable toxic spans—thereby balancing efficiency with interpretability.

\begin{table*}[!htbp]
\small
\centering
\caption{Performance comparison between ToxiCR and ToxiSpanSE. Data is from~\citet{sarker2023automated,sarker2023toxispanse}.}
\label{tab:spannospan}
\resizebox{\textwidth}{!}{%
\begin{tabular}{@{}lllrrrr@{}}
\toprule
\textbf{Tool} & \textbf{Task} & \textbf{Class} & \textbf{Precision} & \textbf{Recall} & \textbf{F1-Score}  \\
\midrule
ToxiCR     & Comment-level binary  & Toxic        & 0.907 & 0.874 & 0.889  \\
           &                      & Non-Toxic   & 0.970 & 0.978 & 0.974  \\
ToxiSpanSE & Token-level span det. & Toxic words      & 0.870 & 0.890 & 0.860     \\
           &                      & Non-Toxic words  & 0.950 & 0.950 & 0.950     \\
\bottomrule
\end{tabular}%
}
\end{table*}

Recent work has also explored multiclass classification to capture different levels of toxicity within a single input, offering an alternative to the conventional binary approach. For instance, one study focusing on maternal insults and stereotyping classified texts into multiple categories—such as “contains maternal insult”, “contains slight maternal insult”, “hard to say”, or “does not contain maternal insult”—and reported higher accuracy compared to binary classification~\citep{sultana2024exploring}. However, the decision boundaries of these multiclass models are not entirely transparent. Moreover, settings such as the model's temperature can also impact the results, leading to variability even when using the same prompt.

The choice of task granularity directly affects human review effort. Whereas binary tags force reviewers to reread the entire article to find the problem, span tags allow them to jump directly to the offending passage. Several user studies with reviewers have shown that better interpretability leads to improved efficiency and accuracy~\citep{xiang2021toxccin, song2023modsandbox}.
However, currently there are no controlled user studies in software engineering to quantify how span-level, sentence-level, or binary outputs affect reviewer efficiency and decision quality at scale. Therefore, future work could include user comparison studies to measure average processing time per review in different output formats, decision accuracy with and without model support, cognitive burden, and trust in the system.

Beyond detection, several studies have investigated generative mitigation strategies. For example, a Conflict Reduction System (CRS) has been proposed that not only detects offensive language but also generates rephrased alternatives to reduce hostility in communities such as GitHub, Gitter, Slack, and Stack Overflow~\citep{cheriyan2021towards}. Similarly, the DeMod tool leverages explainable AI to provide fine‑grained detection results along with personalized content modification suggestions, thereby helping users mitigate toxicity before posting~\citep{li2024demod}. Despite its innovative nature, this approach remains in its early stages and requires further empirical validation on diverse and representative datasets.

\section{Future directions}\label{future}
Future research in toxicity detection could focus on several key areas: enhancing pre-processing techniques, refining detection methods with contextual understanding, improving model interpretability and explainability, and critically addressing ethical governance and policy implications. The following sections outline opportunities within these domains to guide ongoing research.

\subsection{Context-preserving pre-processing}
A potential avenue for future work lies in refining data cleaning techniques to better preserve nuanced semantic meanings in text. For instance, while standard normalization processes often reduce elongated words such as ``loooove" to their base form ``love", this approach may overlook subtle variations in emotional intensity or intent, potentially leading to the loss of valuable contextual or toxicity-related information. Similarly, symbol removal practices might benefit from greater specificity; instead of blanket removal, future research could explore methods to distinguish between symbols that contribute to textual semantics and those that introduce noise. This could involve developing symbol-aware tokenizers that apply language-specific heuristics, such as regex-based detection of emoticons, prior to general symbol removal. Such an approach enables the nuanced preservation of sentiment-conveying symbols, which can then be effectively integrated into analysis, as demonstrated by methods like the blending approach highlighted by~\citep{zou2022sentiment} for enhancing semantic understanding. Developing more granular pre-processing strategies that account for these subtleties could enhance the ability of toxicity detection models to capture contextually rich and emotionally complex patterns, ultimately improving their accuracy and robustness.

\subsection{Context-aware detection methods}
A major challenge in toxicity detection is accurately interpreting context, as certain phrases may be toxic in one setting but neutral or even positive in another. Future work could explore incorporating external knowledge sources, conversation history, or user intent modeling to enhance contextual understanding. For example, leveraging transformer-based models with multi-turn dialogue context could help distinguish between sarcasm, humor, and genuinely harmful content. A recent research~\citep{anuchitanukul2022revisiting} demonstrates that human perception of toxicity is highly dependent on preceding dialogue, and integrating this context into detection models significantly improves accuracy, highlighting the importance of conversational context in toxicity detection.~\citet{zhu2021topic} present a framework that integrates topic modeling with external knowledge bases, offering a potential direction for improving toxicity classification by leveraging similar techniques to understand user intent and conversational context. Future research could build upon these findings by combining multi-turn dialogue modeling, knowledge-aware transformers, and socio-linguistic analysis to develop more nuanced toxicity detection systems capable of differentiating between targeted toxicity and benign discussions.

\subsection{Model interpretability and explainability}
Many toxicity detection models function as black boxes, making it challenging for users and moderators to understand why content is flagged. To improve transparency, future research should prioritize explainable AI (xAI) techniques, such as attention visualization and SHAP (SHapley Additive exPlanations)~\citep{lundberg2017unified}. SHAP assigns contribution scores to individual features, helping to identify which words or phrases influence a model’s toxicity predictions. By quantifying these contributions, it enhances interpretability and allow researchers to assess whether model decisions align with human judgment. Recent models increasingly emphasize interpretability; for example, ToxiSpanSE~\citep{sarker2023toxispanse} improves transparency by detecting toxic spans in code review comments and highlighting specific toxic segments. Additionally, human-in-the-loop approaches that enable users to contest or refine model decisions could foster trust and usability while mitigating unintended consequences, such as excessive or insufficient content flagging.

\subsection{Ethical and Policy Considerations}
Beyond technical advancements, future directions must critically address the ethical governance~\citep{olsen2024right} and policy implications of deploying toxicity detection systems in real-world SE platforms. This includes investigating the development of clear community moderation guidelines that align with platform values and user expectations, ensuring transparency in automated decision-making~\citep{tiwari2023explainable}. Research should also explore robust user feedback mechanisms, allowing individuals to contest classifications and contribute to the refinement of detection models~\citep{molina2022ai}. Furthermore, it is crucial to analyze the broader policy frameworks surrounding content moderation, user consent, and data privacy to ensure that technical enhancements are implemented responsibly and uphold user rights and ethical standards~\citep{llanso2020artificial}. This holistic approach is essential for fostering trust and ensuring the responsible application of toxicity detection technologies.

\section{Ablation study}\label{ablation}
To measure the individual impact of prompting strategies and model choices on SE discussions, we conducted an ablation study with two experiments. The first focused on toxicity detection (classifying code review comments), and the second focused on toxicity mitigation (rewriting individual toxic comments).

\subsection{Detection experiment}
To empirically evaluate the feasibility of leveraging LLMs for toxicity detection in SE communities, we conducted experiments using a dataset of 19,651 labeled code review comments. Each issue was annotated as either toxic or non-toxic. About 19\% of the comments are labeled as toxic. This dataset was created and shared by ~\citet{sarker2023automated}. In the original study, the authors developed a supervised learning based tool to identify toxic code review comments.

The code review comments were collected from FOSS project code reviews and Gitter chat through a four-step process: data mining, stratified sampling, manual labeling, dataset aggregation. We chose three LLMs: GPT-4.1, Gemini-1.5-flash and deepseek-chat. We selected these models due to robust API support and documentation from their providers. Both Google, OpenAI and DeepSeek provide convenient API interfaces and detailed documentation. Among the OpenAI models,  GPT-4.1 understands complex instructions and its I/O cost ($2/$8 per million tokens) is 26\% lower than GPT-4o \citep{openai2024gpt41}. The deepseek-chat model is selected due to its open-source nature and ability to be deployed on local infrastructure, a feature that is particularly valuable for organizations concerned with data privacy or operational costs. Our goal is to assess whether it can deliver toxicity detection performance on par with more resource-intensive proprietary models.All LLMs are evaluated under two prompting conditions. In the few-shot setting, we selected three randomly chosen toxic examples and three non-toxic examples from a different dataset built by \citet{raman2020stress} to maintain independence. In the zero-shot setting, only a textual definition of toxicity was provided without any examples. After obtaining each model’s predictions (“Yes” for Toxic and “No” for Non-Toxic), we computed classification metrics, including accuracy, precision, recall, and F1-score for both classes, as well as overall accuracy.

\subsection{Mitigation experiment}
\begin{figure}[htp]
    \centering
    \includegraphics[width=13cm]{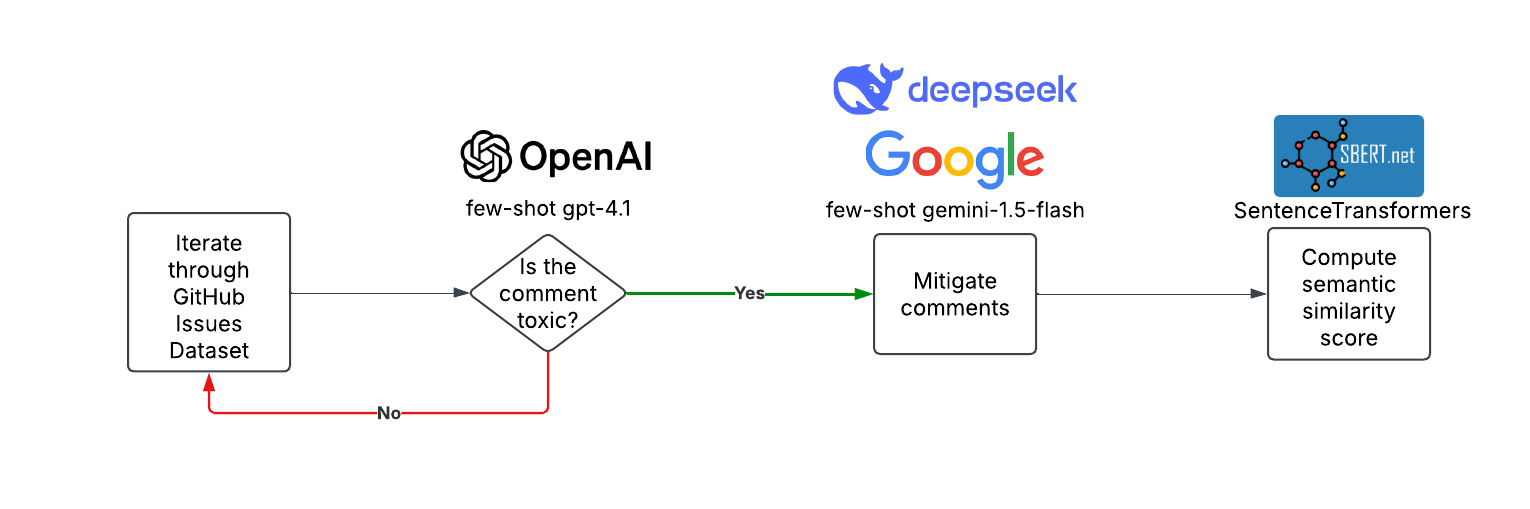}
    \caption{Experimental workflow for toxicity detection and mitigation.}
    \label{fig:Mitigation_Experiment_Workflow}
\end{figure}

In the mitigation experiment, we continued utilizing the dataset from the same source. The key difference from the detection experiment is that, instead of entire issues, the input consisted of individual comments. This granularity enables a more precise evaluation of the mitigation effects~\citep{raman2020stress}. As illustrated in Figure~\ref{fig:Mitigation_Experiment_Workflow}, we employed a structured approach to ensure model independence by utilizing four distinct models, preventing over-reliance on any single approach and enabling a more robust mitigation strategy. Our pipeline first invokes a few-shot GPT-4.1, selected due to its exceptional performance in the previous detection experiment, to identify if a comment is toxic. Prior studies have shown that large-scale language models such as GPT-4 outperform traditional classifiers in zero-shot and few-shot settings, particularly in identifying offensive and harmful language while maintaining high recall rates~\citep{shafikuzzaman2024empirical}. The ChatGPT series of models has been widely recognized for its effectiveness in handling nuanced toxicity detection, especially in software engineering discussions~\citep{mishra2024exploring}. 

Then, the comment will be sent to the Gemini-1.5-flash model and deepseek-chat model to rewrite the text as it has demonstrated strong performance in text generation while preserving semantic consistency. Unlike rule-based text mitigation methods, generative models like Gemini-1.5-flash and deepseek-chat model provide more contextually appropriate rephrasing, ensuring that the rewritten text remains meaningful within the software engineering domain while effectively reducing toxicity~\citep{sultana2024exploring}.

Finally, to evaluate the effectiveness of the rewrites, we measured the semantic similarity using Sentence-BERT, a model designed for capturing sentence-level meaning beyond surface-level wording~\citep{shafikuzzaman2024empirical}. Sentence-BERT was selected due to its superior ability to assess the semantic consistency of rewritten comments compared to classical cosine similarity approaches, making it particularly effective for our mitigation pipeline~\citep{raman2020stress}. This approach provided a robust estimation of how well the mitigated comments preserved the intent of the initial submissions rather than merely matching lexical choices.

\begin{figure}[htp]
    \centering
    \includegraphics[width=12cm]{ 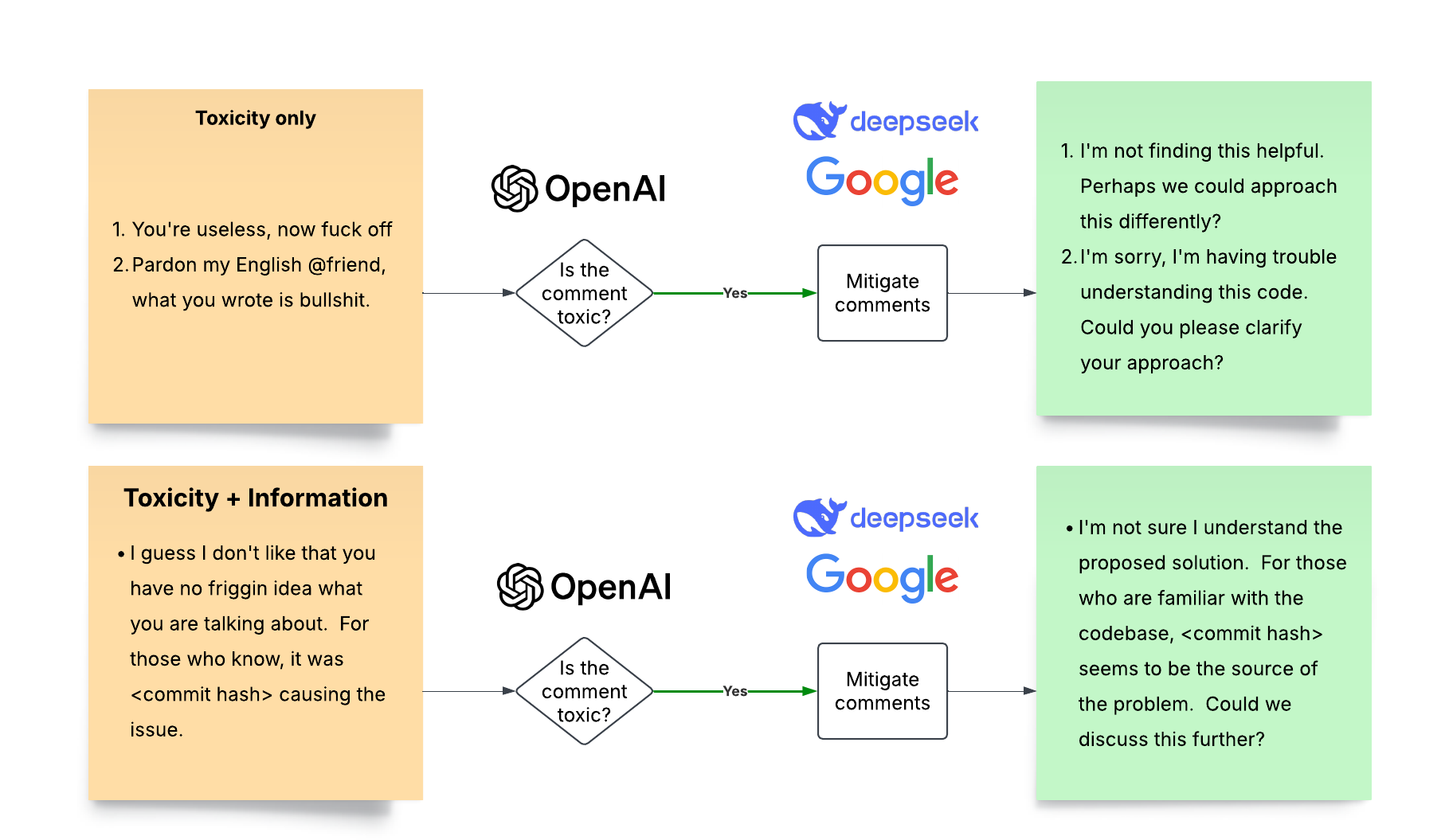}
    \caption{Contrived demonstration of  inputs and outputs from our mitigation pipeline.}
    \label{fig:demo_toxicity}
\end{figure}

\subsection{Prompt and hyper-parameter specification} To remove any ambiguity about reproduction, we used the following prompt wording, model releases, and decoding settings used in every experiment.

\textbf{Zero-shot prompt}: \textit{You are a toxicity detection assistant.
In this task, you will read a code review comment.}
If any part contains rude, offensive, or disrespectful language,
respond with 'Yes'. Otherwise, 'No'.
Comment: \textless comment\textgreater

The few-shot prompt used the same introductory lines, then appended six fixed exemplars—three toxic and three non-toxic—sampled with random state of 42. The exemplar block, is stored in a CSV file once generated, is identical for every run, after which the line Comment: {comment} inserts the test instance.

All calls were issued via the official APIs: OpenAI GPT-4.1, Google Gemini-1.5-flash and DeepSeek deepseek-chat. We did not use stop sequences, presence or
frequency penalties, or logit bias, and each request returned a single
completion (n = 1). Table~\ref{tab:decoding} lists the full
decoding settings.

\begin{table*}[!htbp]
\small
\centering
\caption{Hyper-parameter specification of LLM variants evaluated in experiments. }
\label{tab:decoding}
\resizebox{\textwidth}{!}{%
\begin{tabular}{@{}lcccccc@{}}
\toprule
\textbf{Task} & \textbf{Model} & \textbf{Temperature} & \textbf{Top-p} & \textbf{Top-k} & \textbf{Max tokens} & \textbf{Other} \\
\midrule
Detection & GPT-4.1 & 1.0 & 1.0 & —  & 2\,048 & — \\
Detection & Gemini-1.5-flash & 1.0 & 1.0 & 40 & 2\,048 & — \\
Detection & deepseek-chat & 1.0 & 1.0 & 50 & 4\,096 & 128K context length \\
Mitigation & Gemini-1.5-flash & 1.0 & 1.0 & 40 & 2\,048 & — \\
Mitigation & deepseek-chat & 1.0 & 1.0 & 50 & 4\,096 & 128K context length \\
Semantic similarity & SBERT (all-mpnet-base-v2) & n/a & n/a & n/a & 512 & 768-d embeddings \\
\bottomrule
\end{tabular}%
}
\end{table*}

\section{Results}
This section summarizes the experimental outcomes for toxicity detection and mitigation tasks. We first present results for toxicity detection, highlighting model performance across different prompting methods, followed by detailed findings from our mitigation experiments.

\subsection{Toxicity detection performance}
The performance metrics for the toxicity detection task across all code review comments are summarized in Table~\ref{tab:Evaluation}. Among zero-shot configurations, Gemini-1.5-flash zero-shot achieved the highest overall accuracy (0.85), closely followed by GPT-4.1 zero-shot (0.82). DeepSeek-chat zero-shot performed comparably accuracy(0.84) and showed the highest Toxic recall (0.54) and F1-score (0.56) among all zero-shot models, indicating stronger sensitivity to detecting toxic comments. All three zero-shot models All three models demonstrated strong Non-Toxic detection, with $precision \ge 0.83$ and $recall \ge 0.91$. 

In contrast, few-shot prompting yields mixed results: GPT-4.1 few-shot reached 0.78 accuracy but suffered a drop in Toxic precision (0.40) with no improvement in Toxic recall (0.20) compared to zero-shot. Gemini few-shot improved Toxic recall to 0.55 (from 0.40) but at the expense of precision (0.33) and Non-Toxic recall (0.72), resulting in lower overall accuracy (0.69). DeepSeek few-shot similarly did not benefit from example-based prompting: while maintaining accuracy (0.83), Toxic recall declined to 0.42 and the corresponding F1-score dropped to 0.49.

\begin{table*}[!htbp]
\small
\centering
\caption{Evaluation of prompting methods in toxicity detection (full dataset)}
\resizebox{\textwidth}{!}{%
\begin{tabular}{@{}lccccccc@{}}
\toprule
\textbf{Model \& Prompt}              & \textbf{Accuracy} & \textbf{Non-Toxic P} & \textbf{Non-Toxic R} & \textbf{Non-Toxic F1} & \textbf{Toxic P} & \textbf{Toxic R} & \textbf{Toxic F1} \\
\midrule
GPT-4.1 zero-shot                     & 0.82 & 0.83 & 0.97 & 0.90 & 0.67 & 0.20 & 0.31 \\
GPT-4.1 few-shot                      & 0.78 & 0.82 & 0.93 & 0.87 & 0.40 & 0.20 & 0.27 \\
Gemini-1.5-flash zero-shot            & 0.85 & 0.87 & 0.96 & 0.91 & 0.73 & 0.40 & 0.52 \\
Gemini-1.5-flash few-shot             & 0.69 & 0.87 & 0.72 & 0.79 & 0.33 & 0.55 & 0.42 \\
DeepSeek-chat zero-shot               & 0.84 & 0.91 & 0.91 & 0.91 & 0.59 & 0.54 & 0.56 \\
DeepSeek-chat few-shot                & 0.83 & 0.93 & 0.90 & 0.91 & 0.60 & 0.42 & 0.49 \\
\bottomrule
\end{tabular}%
}
\label{tab:Evaluation}
\end{table*}

\begin{table*}[!htbp]
\small
\centering
\caption{McNemar’s test results for zero-shot vs. few-shot prompting across models.}
\begin{tabular}{lcccc}
\toprule
Model                           & \(b\) & \(c\) & \(\chi^2\) & \(p\)-value \\
\midrule
GPT-4.1                         & 447   & 331   & 16.999     & \(<0.001\)  \\
Gemini-1.5-flash                & 2321  & 634   & 961.961    & \(<0.001\)  \\
DeepSeek-chat (V3.2-Exp)        & 797   & 718   & 4.016      & 0.045       \\
\bottomrule
\end{tabular}
\label{tab:McNemarTest}
\end{table*}

To assess the statistical significance of differences between zero-shot and few-shot prompting, we conducted McNemar’s test on the paired predictions of each model. Table~\ref{tab:McNemarTest} reports the off-diagonal counts (\(b\): zero-shot correct, few-shot incorrect; \(c\): zero-shot incorrect, few-shot correct), the test statistic, and the associated \(p\)-value. For both GPT-4.1 (\(\chi^2=16.999\), \(p<0.001\)) and Gemini (\(\chi^2=961.961\), \(p<0.001\)), the results indicate highly significant differences in overall error rates between zero-shot and few-shot settings. deepseek-chat also exhibits a statistically significant difference between prompting modes ($\chi^2 = 4.016$, $p = 0.045$), but the effect size is much smaller, suggesting that DeepSeek’s detection performance is comparatively less sensitive to prompting strategy and that its strongest performance can be achieved without example-based conditioning.

Crucially, when comparing our results to supervised models, our best zero-shot LLM configuration (Gemini zero-shot) attains only 0.85 accuracy and a 0.50 Toxic F\textsubscript{1} score. This is substantially outperformed by~\citet{sarker2023automated}'s supervised ToxiCR approach, which achieved 0.96 accuracy and an 0.89 Toxic F\textsubscript{1} score on 19,651 code review comments. This suggests that domain-specific, supervised models still substantially outperform zero/few-shot LLM detectors on SE toxicity detection.

\subsection{Toxicity mitigation performance}
The mitigation experiment was applied to all comments initially flagged as toxic. Both Gemini-1.5-flash and deepseek-chat demonstrated strong mitigation capability, with overall success rates exceeding 99.5\%. Gemini achieved a slightly higher Mitigation Success Rate (99.92\%) and fewer failure cases (3 out of 3,547) compared to DeepSeek (99.52\%, 7 failures). However, the two models exhibited different characteristics in terms of semantic preservation. While the average number of rewriting attempts was nearly identical for both models (1.02), DeepSeek produced rewritten comments with higher semantic similarity to the original text (mean similarity: 0.6251 vs. 0.5874 for Gemini), and also yielded a higher proportion of high-similarity rewrites (17.17\% of outputs with similarity $>0.9$, compared to 9.13\% for Gemini). This indicates that DeepSeek is more effective at maintaining the author’s original intent and technical nuance during toxicity reduction.

Fig.~\ref{fig:demo_toxicity} presents representative examples of mitigated comments. In cases where similarity scores are lower for either model, we observe that the original comments often contained little meaningful content beyond their toxic phrasing. In such instances, preserving semantic equivalence is inherently more difficult because removing toxicity necessarily eliminates the primary information conveyed. Overall, these results suggest that while Gemini provides a more conservative and uniformly successful mitigation strategy, DeepSeek better preserves the communicative clarity and technical specificity that are especially important in code review contexts.

\subsection{Study limitations}
Internally, our analysis is constrained by the specific experimental characteristics of the ablation study. First, the limited sample size used in the LLM experiments restricts the statistical power of the analysis and may not fully capture the complete diversity of toxic expressions found in real-world SE communications. An analysis of cases where every LLM failed to detect toxicity reflects the inherent limitations of binary, generalized detectors when applied to developer communications: these include mild or implicit toxicity (e.g., “yeah that sucked, fixed. Done.”), judgmental critiques embedded in technical context (e.g., “function keyword is a bashism and IMHO looks ugly”), and the use of colloquial or informal language with mild profanity (e.g., "confused the hell out of me"). Furthermore, the absence of extensive hyperparameter tuning and reliance on a single epoch of training for the fine-tuned supervised models mean that the performance differences reported may not represent the absolute optimal capacity of those supervised baselines. Similarly, the absence of k-fold cross-validation hinders our ability to fully assess the stability and robustness of the LLM and baseline model predictions across different partitions of the data. Readers should interpret the reported accuracy and F1 scores within the context of these resource and experimental design limitations.

Externally, our review is primarily constrained by a pervasive issue in SE toxicity research: the overwhelming availability of English-language datasets. Given that software development is an inherently global and multilingual process, relying exclusively on English data limits the broader applicability of these findings to development teams communicating in languages such as Chinese, Spanish, or Arabic. This linguistic bias means the identified model failures and mitigation successes may not translate effectively to non-English SE contexts, underscoring a necessary direction for future corpus expansion.

\section{Next steps}
The results of our ablation study demonstrate the potential of LLMs for both toxicity detection and mitigation. While existing models such as GPT-4 and Gemini have shown strong performance, particularly in non-toxic comment classification and mitigation success rates, there are several directions for further research to improve accuracy and applicability.

\begin{itemize}

    \item {\bf Continuous toxicity spectrum}:
    We envision replacing the current binary classification with a continuous toxicity spectrum, assigning each comment a real-valued score between 0 and 1. This fine-grained approach would allow us to detect and quantify low-intensity toxic content that falls below rigid binary thresholds, addressing the issue of mild or implicit toxicity missed by current models. However, moving to a continuum demands more subjective annotation guidelines and requires careful calibration and validation to ensure consistency and reliability across evaluators.

    \item {\bf Domain-specific model training}:
    One key area for improvement is training LLMs on software engineering discourse. Prior research shows general-purpose models struggle with domain-specific terminology and context, leading to misclassifications~\citep{sarker2022built}. Fine-tuning models using labeled SE datasets could enhance their ability to detect nuanced toxicity in developer communications. Beyond supervised fine-tuning, future work should explore complementary strategies such as semi-supervised learning~\citep{zhu2005semi}, leveraging unlabeled SE comments for pretraining to capture broader domain language patterns. Additionally, data augmentation techniques, like back-translation~\citep{edunov2018understanding} of rare toxic examples, could address challenges posed by low-resource toxicity classes, enriching training data and improving model robustness.

    \item {\bf Optimized prompt engineering}:
    Our results suggest that prompt design significantly impacts the effectiveness of LLMs in both detection and mitigation. Previous studies have demonstrated that structured prompts can improve precision and recall in toxicity detection, as prompt clarity, specificity, and framing affect model performance~\citep{mishra2024exploring}. Future work should explore advanced prompt engineering techniques, such as few-shot learning and chain-of-thought prompting, to optimize model outputs and minimize false positives or false negatives.

    \item {\bf Explainable toxicity detection}:
    While current approaches achieve high accuracy, they often lack interpretability, which is crucial in SE settings where developers require actionable insights. Research has highlighted the importance of integrating explanation mechanisms into LLM-based toxicity classifiers, such as providing rationale generation or highlighting specific toxic phrases~\citep{mishra2024exploring}. Future work should investigate explainability frameworks that enhance model trust and usability while ensuring transparency in classification decisions.

    \item {\bf Phased Research Roadmap}:
    We propose a phased research roadmap to guide the iterative implementation and evaluation of these advancements:
    
    \begin{itemize}
        \item Phase 1: Corpus Expansion and Annotation (Estimated: 2-4 months). Focus on collecting and rigorously annotating a larger, diverse software engineering toxicity corpus from various SE platforms, capturing domain-specific nuances and low-resource toxicity classes, including developing guidelines for the continuous toxicity spectrum.
    
        \item Phase 2: Model Adaptation and Optimization (Estimated: 1-2 months). Build upon the expanded corpus by fine-tuning existing LLMs or developing new domain-specific architectures. This involves exploring training strategies like semi-supervised learning and advanced prompt engineering to optimize model performance.
    
        \item Phase 3: Explainable Prototype Deployment and User Evaluation (Estimated: 1-2 months). Deploy an explainable prototype toxicity detection system in a controlled SE environment, integrating explainability frameworks for actionable insights. Refine model performance using Reinforcement Learning from Human Feedback (RLHF,~\citet{christiano2017deep}) based on user preferences and context. Conduct thorough user evaluations to gather feedback, assess usability, and refine the system for real-world applicability.
    \end{itemize}

\end{itemize}

\section{Conclusion}
As LLMs become increasingly integrated across diverse sectors—ranging from education and healthcare to software engineering—the challenge of managing toxic language remains ever more pressing. In this meta-review, we have synthesized recent advances in toxicity detection and mitigation, drawing on a wide spectrum of studies that span dataset construction, annotation protocols, and pre-processing techniques. This work echoes our introduction by highlighting the dual-edged nature of LLMs: while they offer robust contextual understanding and generative capabilities that can uncover nuanced toxic expressions, they also risk propagating harmful content if not carefully managed.

Our findings indicate that the efficacy of toxicity detection is critically dependent on the quality of datasets and the consistency of annotation standards, with pre-processing and sampling strategies playing a pivotal role in shaping model performance. Conventional detection methods, although effective in identifying overt toxicity, often fall short in capturing subtle, context-dependent expressions—an area where LLMs demonstrate clear advantages. Additionally, generative rewriting approaches show promise in mitigating toxicity by reformulating harmful content while largely preserving the original meaning, thus fostering more constructive communication.

Beyond technical performance, the implications of improved toxicity detection and mitigation in SE extend significantly to organizational culture~\citep{bhat2021say} and developer well-being~\citep{sarker2022identification}. By fostering healthier, more inclusive open-source communities, these advancements can enhance collaboration, improve developer retention, and support a more equitable and welcoming environment for all contributors. Ultimately, reducing toxic interactions can lead to more productive and sustainable development ecosystems.

In summary, this paper not only provides a comprehensive overview of current methodologies for combating toxic language but also underscores existing limitations and the need for further refinement. Future research should focus on domain-specific model tuning, advanced prompt engineering, and enhanced interpretability to build more transparent toxicity mitigation systems. Through these efforts, we hope to pave the way for safer and more inclusive online environments.

\clearpage
\backmatter
\bibliography{main}

\end{document}